\def\year{2022}\relax
\documentclass[letterpaper]{article} 

\usepackage{amsmath}
\usepackage{amssymb}
\usepackage{setspace}
\usepackage{multirow} 
\usepackage{booktabs}
\usepackage{subfigure}
\usepackage{tabularx}
\usepackage{verbatim}

\usepackage{aaai22}  
\usepackage{times}  
\usepackage{helvet}  
\usepackage{courier}  
\usepackage[hyphens]{url}  
\usepackage{graphicx} 
\usepackage{natbib}  
\usepackage{caption} 
\DeclareCaptionStyle{ruled}{labelfont=normalfont,labelsep=colon,strut=off} 
\usepackage{algorithm}
\usepackage{algorithmic}

\usepackage{newfloat}
\usepackage{listings}
\floatstyle{ruled}
\newfloat{listing}{tb}{lst}{}
\floatname{listing}{Listing}
\title{Alleviating Mode Collapse in GAN via Pluggable Diversity Penalty Module}
\author{
 Sen Pei\textsuperscript{\rm 1}, \quad Richard Yi Da Xu\textsuperscript{\rm 2},\quad Shiming Xiang\textsuperscript{\rm 1}, \quad Gaofeng Meng\textsuperscript{\rm 1}
}
\affiliations{
    \textsuperscript{\rm 1}Institute of Automation, Chinese Academy of Sciences\\
    \textsuperscript{\rm 2}University of Technology Sydney\\

    {\tt\small peisen2020@ia.ac.cn, yida.xu@uts.edu.au, \{smxiang, gfmeng\}@nlpr.ia.ac.cn}
}

\usepackage{bibentry}

\begin{document}

\maketitle

\begin{abstract}
The vanilla GAN suffers from mode collapse deeply. This problem usually manifests as that the images generated by generators tend to have high similarity amongst them, even though their corresponding latent vectors have been very different. In this paper, a Pluggable Diversity Penalty Module (PDPM) is proposed to address this issue. The motivation behind our framework is to enforce the generator to generate images with distinct features if their corresponding latent vectors are different. To this end, PDPM extracts the feature maps of fake images via discriminator first, and then, the normalized Gram matrix is used to measure the similarity of these feature maps. In latent space, the similarity of latent vector pairs is given in the same way for consistency. With these similarity relationships among latent vectors and their corresponding image features, PDPM penalizes the generator if two latent vectors with low similarity are mapped to fake images with similar features. This will reduce the chance of mode collapse in GAN significantly. Further, the proposed PDPM has been extensively compared with some mainstream methods such as ALI, DCGAN, Unrolled GAN, WGAN\_GP, VEEGAN, PacGAN, BourGAN, StarGAN and MSGAN on several public datasets, both visual and quantitative results show that PDPM achieves SOTA performance in image generation, image data augmentation, domain translation and other tasks.
\end{abstract}

\section{Introduction}
In the past few years, GAN \cite{5} has been widely used in image generation, image inpainting, style transfer and super-resolution reconstruction, etc. However, with the great progresses in GAN, an essential problem has always been with us, that's mode collapse. This phenomenon heavily harms the diversity and quality of images generated by generator. In this paper, we mainly focus on mode collapse alleviation and aim to generate data in high diversity based on the available, and further, apply the augmentated data into downstream tasks for better performance.

\begin{figure}[t]
\begin{center}
   \includegraphics[width=1.0\linewidth]{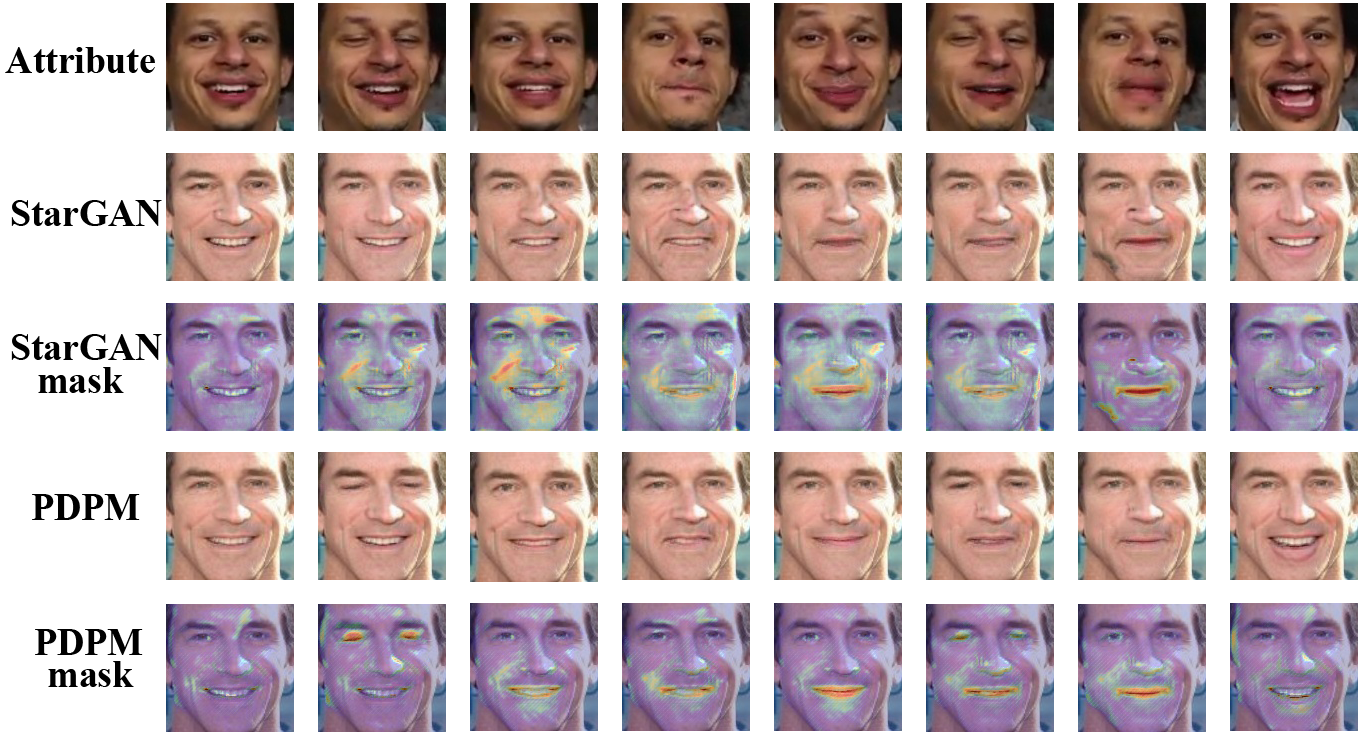}
\end{center}
\caption{\textbf{Domain translation with PDPM.} Both StarGAN and PDPM are trained 10 epochs. PDPM can transfer the facial expressions of attribute images to other faces while the vanilla StarGAN can not capture the change of eyes, this indicates PDPM converges much faster. Besides, it is clear to see that PDPM generates accurate attention masks of changes compared with the first column while StarGAN captures much more background.}
\label{fig:animation}
\end{figure}

In general, mode collapse usually manifests as that the trained generator can only generate images in some specific classes which really harms the data diversity. Currently, to the best of our knowledge, there are two main ways to alleviate mode collapse, modifying the architecture (or training method) of GANs or refining the loss function. The main drawback of the former is its poor generalization performance since it is effective just for some specific networks, for example, in Unrolled GAN \cite{25}, the generator has to consider both its current state and the state of discriminator after $K$ iterations which is hard to apply to other models. By contrast, the latter method usually has better generalization ability, but it is difficult to design an universal module, for example, in DRAGAN \cite{31} and MSGAN \cite{24}, new penalty terms are introduced for improving data diversity, but in our experiments (see Figure \ref{fig:men-women}), we notice these methods may generate some noisy pixels which harms the image quality. Besides, using multiple GANs can alleviate this problem to some extent, but due to its high cost, this method is rarely adopted in practice. Up to this day, most approaches of mode collapse alleviation start with the original data space while few methods deal with this problem via features of the fake images.

Moreover, in our experiments (see Figure~\ref{fig:dp-collapse-sample}), we notice an abnormal phenomenon that sometimes very different latent vectors may be mapped to similar images which is the essential characteristic of mode collapse. Besides, in traditional GANs, the images generated by generator are more like the combination of several images, and this usually leads to low image resolution and quality. In brief, the observations stated above are appearances of mode collapse, and  they indicatie the necessity of addressing this issue.

\begin{figure}[t]
\begin{center}
   \includegraphics[width=1.0\linewidth]{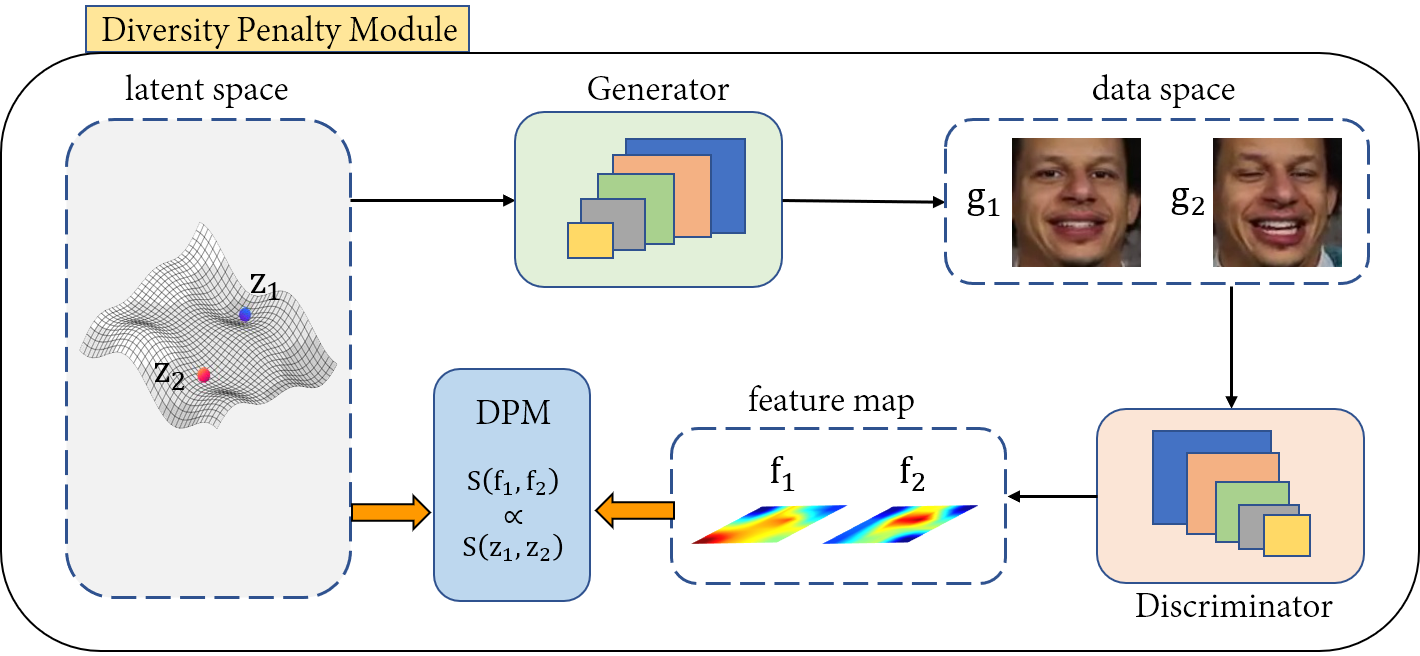}
\end{center}
\caption{\textbf{The proposed PDPM.} In the framework above, $f_1$ and $f_2$ are feature maps extracted from the discriminator, $z_1$ and $z_1$ are latent vectors while $g_1$ and $g_2$ are their corresponding fake images. $S(\cdot)$ indicates the similarity measurement function. The key idea of PDPM is that the similarity relationship of fake images' features should be consistent with their corresponding latent vectors.}
\label{fig:proposed-framework}
\end{figure}

To alleviate the effects of mode collapse while avoiding the drawbacks of previous methods, a novel pluggable diversity penalty module is proposed in this paper, hereinafter, \textbf{PDPM}. Figure~\ref{fig:proposed-framework} shows the pipeline of our framework. Concretely, the more difference between latent vectors the more different their corresponding fake images should be, i.e., if two latent vectors are different, then PDPM enforces generator to generate two images with different features. Unlike current mainstream methods, PDPM performs constraints in feature space which is more robust than that in data space. In latent space, the similarity among latent vectors is given using Gram matrix. However, in data space, each image usually has a great amount of pixels which are uncessary for distinguishing, and in fact, ~\cite{11} find that the feature representations can better describe an image than pixels. Thus, PDPM calculates the similarity of images via their corresponding feature maps. Besides, nonlinear mapping is performed for normalizing the similarity values. The key idea of PDPM  is that the similarity of feature pairs should be consistent with the similarity of their corresponding latent vector pairs. This paper has the following contributions: 
\begin{itemize}
\item A novel block named PDPM is proposed to alleviate mode collapse in GAN. PDPM has better generalization ability compared with most current methods, it can be used in almost all GANs as a pluggable attachment. Besides, PDPM performs constraints in feature space which is more robust and has better pixel value stability;

\item PDPM has great transfer ability and low computation cost. It can be used in image generation, image data augmentation, domain translation and so on, this indicates PDPM is not sensitive to tasks. Besides, PDPM is almost parameters-free which has only one balance coefficient;

\item Compared with other complex methods, PDPM is effective yet easy to perform. The results in Figure~\ref{fig:synthetic} on 2D synthetic dataset show that PDPM can help GAN capture much more modes effectively, and Figure~\ref{fig:animation} also suggests PDPM has good performance in domain translation. In image data augmentation, PDPM introduces a markable accuracy improvement on ResNet. In image generation, PDPM outperforms MSGAN, WGAN\_GP, WGAN\_GP\_MS and some other SOTA architectures both visually and quantitatively (IS and FID).
\end{itemize}

\section{Related Work}

\noindent
\textbf{Mode Collapse Reduction}\quad For improving data diversity and stable training, researchers have done a lot of work. In Unrolled GAN~\cite{25}, Metz \emph{et al.} define the generator objective with respect to an unrolled optimization of the discriminator. The generator has to consider both its current state and the state of discriminator after $K$ iterations. This can lead to a better solution, but it is hard to apply to other models. In Energy-based GAN \cite{32}, Zhao \emph{et al.} use entropy to measure the diversity of images generated by generator while maintaining low energy state. In VEEGAN~\cite{26}, Srivastava \emph{et al.} introduce a variational principle for estimating implicit probability distributions which can help avoide mode collapse. Further, in PacGAN~\cite{28}, Lin \emph{et al.} let the discriminator make decisions based on multiple samples from the same class which can penalize generator with mode collapse. In BourGAN~\cite{29}, Xiao \emph{et al.} treat modes as a geometric structure of data distribution in a metric space which also leads to a better genertor. Recently, in MSGAN~\cite{24}, Mao \emph{et al.} modify the objective function for encouraging the generators to explore more minor modes in data space. By contrast, our proposed PDPM captures modes in feature space which is more robust than MSGAN.\\

\noindent
\textbf{Data Augmentation Learning}\quad Image data augmentation has been proven to be effective in practice. In~\cite{8}, data augmentation is used to reduce overfitting. Also in~\cite{4}, Shorten \emph{et al.} find that even simple techniques such as cropping, rotating and flipping can have markable effects on reducing overfitting. Currently, as \cite{9}, image data augmentation mainly has three branches which are traditional transformations, generative methods and learning the augmentation. The former method has been well studied while the latter has very high computation cost like NAS in \cite{33}. In generative models, GAN is the representative, but it is rarely used due to the limited diversity of generated data caused by mode collapse. That's part of PDPM's motivation.\\

\noindent
\textbf{Convergence and Stability of GANs}\quad The stability of training process and convergence speed are vital for GANs. In~\cite{5}, the vanilla GAN is proposed for generating high-quality images, but at that time, it is not an easy task to train GAN stably due to the imbalance between generator and discriminator. Further, in~\cite{6}, wasserstein distance is used to measure the similarity between distributions instead of KL-divergence, this reduces the difficulty of GAN training greatly, and then, in~\cite{7}, the gradient penalty term is proposed to enforce the Lipschiz constraint instead of using weight clipping as WGAN. These difficulties in training GANs suggest refining loss function is not a trivial task, and indicate the necessity of evaluating convergence and stability of GANs. \\

\noindent
\textbf{Feature Representations of CNN}\quad A deep convolutional layer can extract the feature of an input image accurately. In~\cite{10}, deconvnet is used to visualize the features that a fully trained model has learned. Furthermore, in~\cite{11}, Zhou \emph{et al.} demonstrate that the convolutional neural networks are able to localize the discriminative regions of image. Based on this finding, we use the features extracted from discriminator to represent the images instead of using images directly. And in~\cite{1}, the proposed Grad-CAM method also supports the results in ~\cite{11}. Figure~\ref{fig:grad-cam} in Appendix shows some Grad-CAM results on CelebA~\cite{23} with our trained discriminator, and these results show PDPM can capture image features accurately.\\

\section{Motivation}
\begin{figure}[htpb]
\begin{center}
   \includegraphics[width=1.0\linewidth]{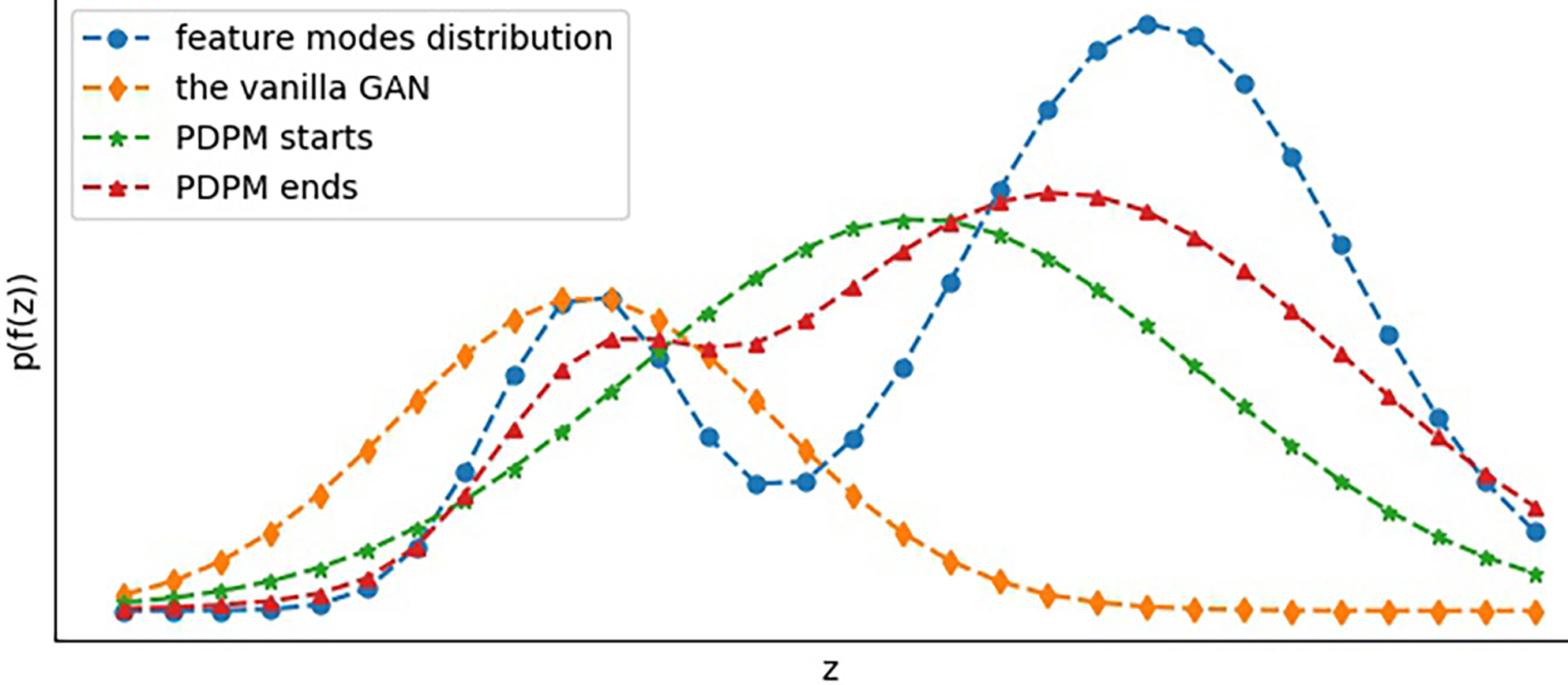}
\end{center}
\caption{\textbf{Illustration of mode collapse in feature space.} The {\tt\small{feature modes distribution}} is the implicit real distribution of data, and {\tt\small{the vanilla GAN}} indicates the modes captured by generator without PDPM.}
\label{fig:mode_collapse}
\end{figure}

To alleviate mode collapse, generator must capture much more modes of the available data. As in Figure \ref{fig:mode_collapse}, $z$ represents latent vectors while $f(z)$ represents the features of corresponding fake image. In feature space, mode collapse manifests as that only a part of modes can be captured by the generator like {\tt\small{the vanilla GAN}} group in Figure \ref{fig:mode_collapse}. And as a result, the data generated by generator will gather in some specific classes or some typical features. Inspired by this, PDPM lets the generator to generate images with more discrete features first, and then, with the development of discriminator, PDPM's penalty term will make the latent vectors distribute around centers of feature modes mainly. The case that different latent vectors clustered around the similar feature by generator will result in higher penalty loss. In brief, PDPM makes the features of fake images much more discrete first ({\tt\small{PDPM starts}} in Figure \ref{fig:mode_collapse}), and then with the help of discriminator and regularization term, to assign these latent vectors to different mode centers ({\tt\small{PDPM ends}} in Figure \ref{fig:mode_collapse}). In next section, the pipeline of PDPM  and its mechanism will be given in detail. 

\section{Pluggable Diversity Penalty Module}
As introduced in previous sections, PDPM penalizes the generator if two different latent vectors are mapped to images with similar features. In this section, $G(\cdot)$ and $D(\cdot)$ are used to indicate generator and discriminator, $p_z$ and $p_r$ are distributions of latent vectors and the real data. Besides, fake images indicate the images generated by generator, and if not specified, $f$ is used to represent the features of fake images extracted from discriminator.

\subsection{Measurement of similarity}
Suppose $p_{z}(z)$ is the distribution of latent vectors which follows a standard normal distribution, a batch vectors $\{z_1, z_2, ..., z_m\}$ are randomly sampled from $p_{z}(z)$, and then, the normalized Gram matrix can be shown as:
\begin{equation}
G_z^*(i, j)=\frac{z_i^Tz_j}{||z_i||_2\cdot ||z_j||_2} \label{eq:original-latent-sim}
\end{equation}
where $\|\cdot\|_2$ represents l2-norm. It is reasonable to suppose that $z_i$ and $z_j$ are independent identically distributed (\emph{i.i.d.}), and in fact, $G_{z}^*$ still follows a gaussian distribution which can be derived from the following claim:\\[6pt]
\emph{$f(x)$ and $g(x)$ are Gaussian PDFs with means $\mu_f$ and $\mu_g$ and standard deviations $\sigma_f$ and $\sigma_g$, then the product of $p(x)$ and $g(x)$ follows a scaled Gaussian distribution with $\mu=\frac{\mu_f\sigma_g^2+\mu_g\sigma_f^2}{\sigma_f^2+\sigma_g^2}$ and $\sigma=\sqrt{\frac{\sigma_f^2\sigma_g^2}{\sigma_f^2+\sigma_g^2}}$. The scale factor is \rm $s=\frac{\sigma_f^2\sigma_g^2}{\sqrt{\sigma_f^2+\sigma_g^2}}\mathbf{exp}\left [ -\frac{(x-\mu)^2}{2\sigma^2} \right ]$.}\\

Likewise, the similarity of feature pairs can be got as follows:
\begin{equation}
G_f^*(i, j)=\frac{f_i^Tf_j}{||f_i||_2\cdot ||f_j||_2} \label{eq:original-feature-sim}
\end{equation}
where $f_i$ represents flattened feature map of the \emph{i-th} fake image extracted from discriminator. Since the values in Eq (\ref{eq:original-latent-sim}) and Eq (\ref{eq:original-feature-sim}) can be zero or negative, performing division directly doesn't make sense, and thus, the \emph{sigmoid} function is used to scale them.
The scale factor is denoted by $s$ and Eqs (\ref{eq:original-latent-sim}) and (\ref{eq:original-feature-sim}) can be revised as:
\begin{equation}
G_z(i, j)=\sigma(s\frac{z_i^Tz_j}{||z_i||_2\cdot ||z_j||_2}) \label{eq:revised-latent-sim}
\end{equation}

\begin{equation}
G_f(i, j)=\sigma(s\frac{f_i^Tf_j}{||f_i||_2\cdot ||f_j||_2}) \label{eq:revised-feature-sim}
\end{equation}

\subsection{Loss function}
For alleviating mode collapse, the diversity penalty module should have the following attributes:

\begin{itemize}
\item if two latent vectors are similar, their corresponding fake images are unlikely to be very different.
\item if two latent vectors are different, their corresponding images have to be different likewise, which means the corresponding feature maps exist much difference.
\end{itemize}

\noindent Obviously, the diversity penalty module should pay much attention to the second situation which often results in mode collapse. Through these observations, the diversity penalty module is designed as follows:
\begin{equation}
DP(z)=\frac{1}{m^2}\sum_{i=1}^{m}\sum_{j=1}^{m}\frac{G_f(i,j)}{G_z(i,j)}\label{eq:dp-z}
\end{equation}
where \emph{m} represents the batch size, and $DP(z)$ has to be minimized when training GANs. Taking the vanilla GAN for example, $D(\cdot)$ is trained to maximize the probability of assigning the correct label to both real images and fake images, also, $G(\cdot)$ is trained simultaneously to get high score from $D(\cdot)$. Thus, the basic loss function of GAN can be formulated as follows:
\begin{align}
&\max_{G}  L_G(z)= \mathbb{E}_{z\sim p_{z}}D(G(z)) \\
&\min_{D} L_D(z, x)=\mathbb{E}_{z\sim p_{z}}D(G(z))-\mathbb{E}_{x\sim p_r}D(x) \\
&\max_{G}\min_{D}\,\mathbb{E}_{z\sim p_{z}}D(G(z))-\mathbb{E}_{x\sim p_r}D(x) 
\end{align}
To perform diversity penalty, we just need to add diversity penalty loss to generator. The loss function of GAN with PDPM can be formulated as follows: 
\begin{align}
&\max_{G}  L_G(z)= \mathbb{E}_{z\sim p_{z}}D(G(z)) - \lambda \mathbb{E}_{z\sim p_{z}}DP(z)\label{eq:gloss}\\
&\min_{D} L_D(z, x)=\mathbb{E}_{z\sim p_{z}}D(G(z))-\mathbb{E}_{x\sim p_r}D(x) \label{eq:dloss}\\
&\max_{G}\min_{D}\,\mathbb{E}_{z\sim p_{z}}D(G(z))-\mathbb{E}_{x\sim p_r}D(x) - \lambda \mathbb{E}_{z\sim p_{z}}DP(z) \label{eq:total-loss}
\end{align}
where $\lambda$ is a balance coefficient of diversity penalty term. The loss function of discriminator remains unchanged. 
According to Eqs (\ref{eq:gloss}), (\ref{eq:dloss}) and (\ref{eq:total-loss}), the training pipeline can be summarized in \textbf{Algorithm 1}.

\subsection{Mechanism Explanations}
When training GANs, the discriminator is usually trained $k$ times while generator is trained only once, and that means discriminator usually converges better than generator. At the begining of training, PDPM enforces generator to generate fake images with discrete features, and this makes it possible for generator to capture more feature modes  like {\tt\small{PDPM starts}} in Figure \ref{fig:mode_collapse}. At that time, the discriminator is not well trained, and it dose not penalize the generator severely. Then, as the discriminator is trained better and better, it will enfoece the generator to map the latent vectors around peaks of the feature distribution like {\tt\small{PDPM ends}} in Figure \ref{fig:mode_collapse}, and the case that latent vectors are mapped to the saddle of feature distribution will get low score from discriminator. Thus, when PDPM converges, most of the latent vectors will be mapped to the surroundings of feature modes while little vectors scatter around untypical feature centers.

\begin{algorithm}[h] 
\label{algorithm-1}
\caption{\small{GAN} with \small{PDPM} training via mini-batch \small{Adam}} 
\begin{algorithmic}[1] 
\FOR{total training \emph{epochs}}
\FOR{\emph{k} times} 
\STATE  Sample a batch data from $p_z$ : $\{z_1, z_2, ...,z_m\}$; 
\STATE  Sample a batch data from $p_r$ : $\{x_1, x_2, ...,x_m\}$; 
\STATE  Update discriminator :
\STATE  \qquad $\theta_d\leftarrow \;\theta_d-\nabla_{\theta_d}\frac{1}{m}\sum_{i=1}^{m}L_D(z_i, x_i)$
\ENDFOR 
\STATE  Sample a batch data from $p_z$ : $\{z_1, z_2, ...,z_m\}$;
\STATE  Update generator :
\STATE  \quad \;\; \qquad $\theta_g\leftarrow \;\theta_g-\nabla_{\theta_g}\frac{1}{m}\sum_{i=1}^{m}L_G(z_i)$
\ENDFOR 
\end{algorithmic} 
\end{algorithm}

\section{Experiments}

In this section, PDPM is evaluated from several different views. First, in {\tt\small{Basic Attribute Evaluation of PDPM}} part, the feasibility analysis of similarity measurement and the convergence performance of PDPM will be talked in detail, and next, in {\tt\small{Ablation Study}} part, both visual and quantitative comparisons between PDPM and other typical architectures such as ALI~\cite{27}, Unrolled GAN~\cite{25}, VEEGAN~\cite{26}, PacGAN~\cite{28} and BourGAN~\cite{29} on 2D Synthetic Datasets will be given, which indicates the efficiency of PDPM. Further, in {\tt\small{Multi Task Applications}}, PDPM is applied into domain translation, image generation, image data augmentation and other tasks, and the results show PDPM outperforms most mainstream GANs such as DCGAN \cite{15}, WGAN\_GP \cite{7} and MSGAN \cite{24}.

\subsection{Datasets}
The datasets used in our experiments are MNIST, Fashion-MNIST, CIFAR-10, CelebA and 2D Synthetic Datasets. For the first three datasets, only training set is used while no changes are made in testing set. In some tables, M, F-M and C10 are used to represent MNIST, Fashion-MNIST and CIFAR-10 respectively. 

\subsection{Training Details}
Unless specified, $\mathbf{Adam}$ optimizer with $\beta_1$=0.5 and $\beta_2$=0.9 is used for training GANs, and $\mathbf{SGD}$ optimizer with weight decay(1e-4) and momentum(0.9) is used for training ResNet. The initial learning rates are 1e-4 and 1e-3 for GANs and ResNet respectively. The traditional data augmentation methods contain {\tt RandomHorizontalFlip, RandomVerticalFlip, RandomResizedCrop, RandomRotation} and {\tt RandomColorJitter}.

\subsection{Basic Attribute Evaluation of PDPM}
\begin{figure}[htpb]
\begin{center}
   \includegraphics[width=0.71\linewidth]{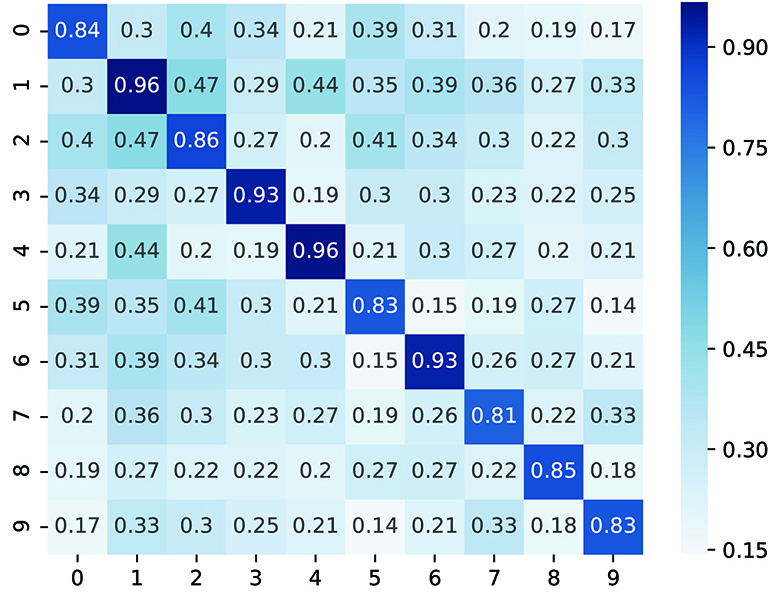}
\end{center}
   \caption{\textbf{Similarity Analysis on Fashion-MNIST.} 
For avoiding occasionality, 5k images are sampled per class. The similarity is computed via Eq (\ref{eq:revised-feature-sim}).} 
\label{fig:similarity-fm}
\end{figure}

Similarity measurement must has two basic characteristics:
\begin{itemize}
\item The similarity value should be higher within classes than between classes.
\item Visually similar images should be close in feature space.
\end{itemize}
Picking Fashion-MNIST images as samples, the similarity values among different categories are calculated via their corresponding feature maps extracted from discriminator. Results shown in Figure~\ref{fig:similarity-fm} tell that the similarity value is higher within classes than between classes.

Further, similar operation is performed within one specific class on Fashion-MNIST to verify the second character stated above. Results are attached in Appendix Figure~\ref{fig:fm-one-class}, they confirm that visually similar images are also similar in feature space and vice versa. These statistical results verify the reasonability of our similarity measurement.

Besides, for GANs, whether it can converge stably or not is vital, and thus, the evaluation of PDPM on MNIST, Fashion-MNIST, CIFAR-10 and CelebA is made, respectively. Architectures of GAN are contained in Appendix Table~\ref{architecture}. Figure \ref{fig:gloss} gives the convergence results of domain translation on CelebA, the detailed results of other datasets are attached in Appendix Figure \ref{fig:convergence}. In domain translation, StarGAN \cite{30} is set as baseline, two groups with PDPM are set for comparison. From the results shown in Figure \ref{fig:gloss}, it is clear to see that PDPM can accelerate the convergence of generator significantly. 

\begin{figure}[htpb]
\begin{center}
    \includegraphics[width=0.45\textwidth]{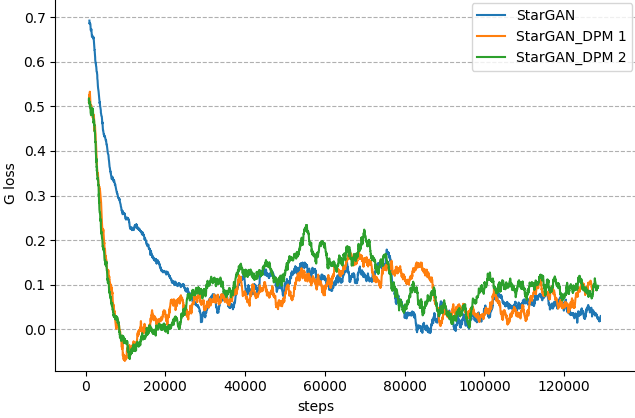}
\end{center}
    \caption{\textbf{Loss of generator.} The balance coefficient $\lambda$ in PDPM is set to 1e-3 in StarGAN\_PDPM 1 and 1e-4 in StarGAN\_PDPM 2.}
    \label{fig:gloss}
\end{figure}

This acceleration is achieved because PDPM can capture accurate feature representations which are vital in facial expression transfer. Besides, in Figure \ref{fig:animation}, the first column indicates the original images and their corresponding facial masks, the following columns are results of facial expression transfer and their corresponding attention masks. These attention masks should capture the changes between the image after transformation and the original. It can be seen that PDPM can generate much clearer facial attention masks with less background which can bring better and smoother detail changes compared with the vanilla StarGAN. Besides, StarGAN in Figure \ref{fig:animation} can not transfer the changes of eyes to new face image, this indicates that when training with same epochs, the vanilla StarGAN converges much worse than PDPM.

\subsection{Ablation Study}
In this part, the effects of PDPM are stated in detail. First, in {\tt\small{Evaluation on Basic Datasets}}, both visual and quantitative results of mode collapse and mode collapse alleviation are given on MNIST, Fashion-MNSIT and CIFAR-10, further, in {\tt\small{Evaluation on 2D Synthetic Datasets}}, the comparison between some typical GANs with and without PDPM are given for precise comparison.

\subsubsection{Evaluation on Basic Datasets}
\begin{figure*}[t]
	\subfigure[WGAN\_GP without PDPM on MNIST]{
        \includegraphics[width=0.49\textwidth]{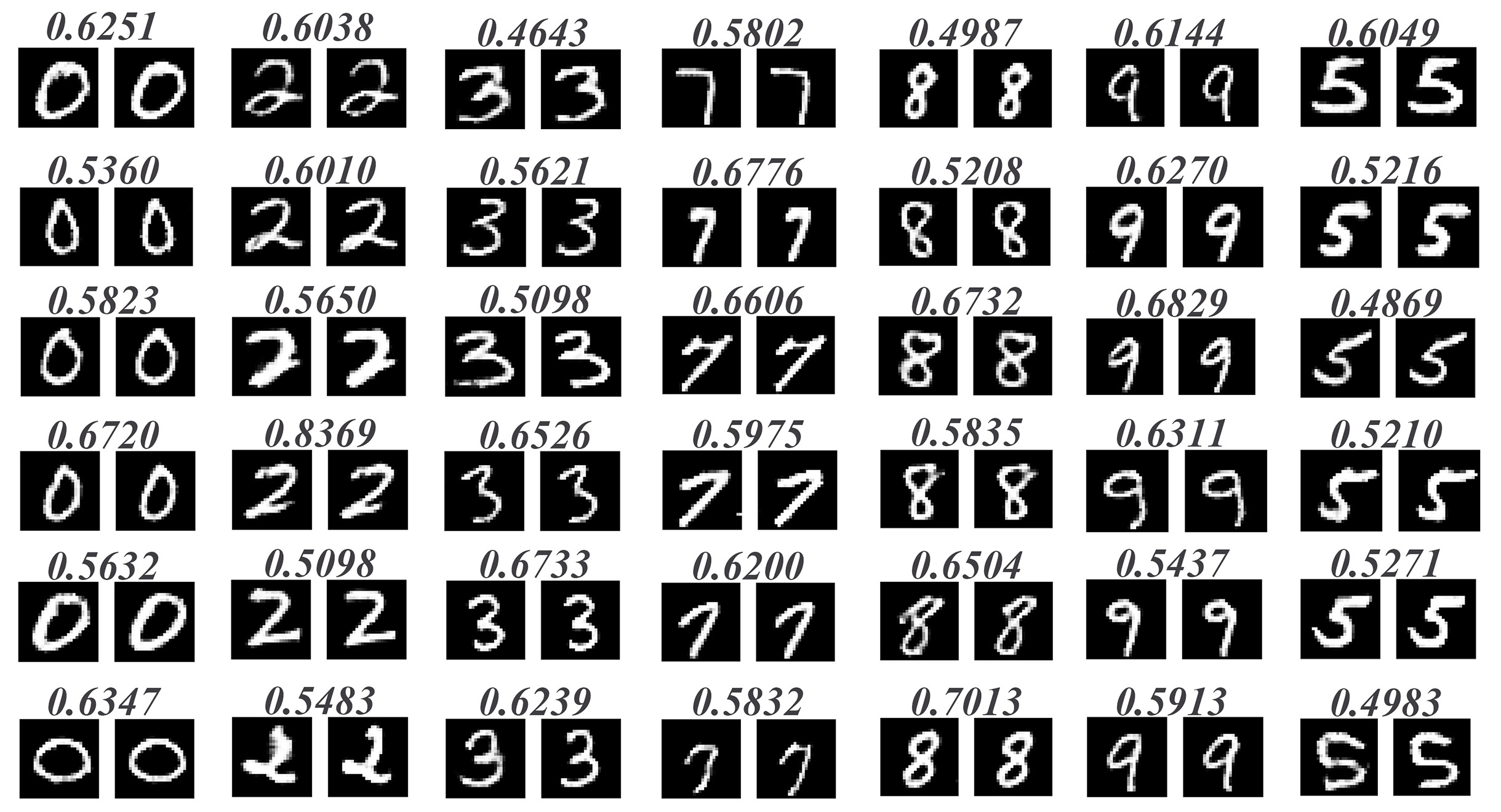}
    }
    \subfigure[WGAN\_GP with PDPM on MNIST]{
        \includegraphics[width=0.49\textwidth]{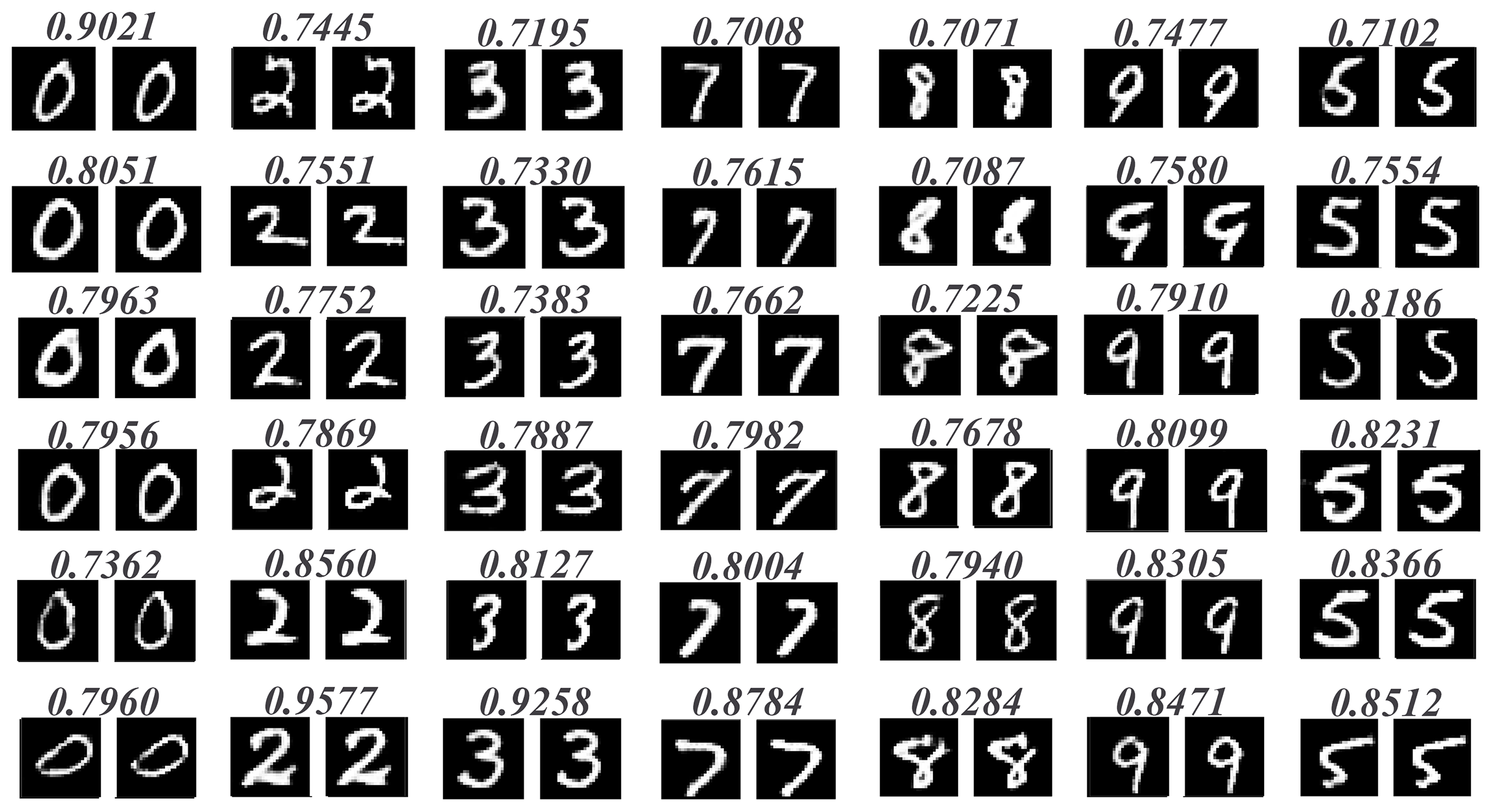}
    }
    \caption{\textbf{Alleviation of mode collapse via PDPM.} (a) WGAN\_GP without PDPM. (b) WGAN\_GP with PDPM $\lambda$=5. The value above each image pair indicates the similarity value of their latent vectors. It can be found that in GAN without PDPM, latent vectors with low similarity value can be mapped to similar images while PDPM not.}
    \label{fig:dp-collapse-sample}
\end{figure*}

In vanilla GANs,the latent vectors even with very low similarity may be mapped to similar images, but with PDPM, this phenomenon is alleviated since this situation will result in higher loss. That is, PDPM makes similar fake images have corresponding latent vectors with higher similarity. Using the method shown in Appendix Figure \ref{fig:mode-collapse-sample}, the similar images and their corresponding latent vectors can be got simultaneously with our trained generator. With these fake images and their latent vectors, the similarity value can be calculated via Eqs (\ref{eq:revised-latent-sim}) and (\ref{eq:revised-feature-sim}). Part of these results are shown in Figure \ref{fig:dp-collapse-sample}, others are attached in Appendix. These results indicate that PDPM prevents the generator from mapping latent vectors with low similarity to similar fake images. 

\begin{table}[htpb]
\setlength{\abovecaptionskip}{0.2cm}
\caption{\textbf{Statistic results of diversity penalty module.}}
\begin{tabular}{c|l|lll}
\hline
\multicolumn{2}{l|}{\small{Dataset}}   & \small{WGAN\_GP}         & \small{PDPM $\lambda$=5}        & \small{PDPM $\lambda$=10}           \\ \hline
\multirow{10}{*}{\small{M // FM}} & 1  & 0.68 // 0.65 & 0.78 // 0.82 & 0.82 // 0.84 \\
                          & 2  & 0.67 // 0.63 & 0.66 // 0.77 & 0.65 // 0.85 \\
                          & 3  & 0.64 // 0.63 & 0.77 // 0.82 & 0.77 // 0.89 \\
                          & 4  & 0.69 // 0.61 & 0.80 // 0.83 & 0.78 // 0.81 \\
                          & 5  & 0.64 // 0.66 & 0.78 // 0.84 & 0.77 // 0.81 \\
                          & 6  & 0.68 // 0.64 & 0.78 // 0.83 & 0.78 // 0.81 \\
                          & 7  & 0.64 // 0.63 & 0.75 // 0.84 & 0.77 // 0.86 \\
                          & 8  & 0.61 // 0.66 & 0.74 // 0.86 & 0.75 // 0.87 \\
                          & 9  & 0.67 // 0.64 & 0.80 // 0.79 & 0.80 // 0.83 \\
                          & 10 & 0.64 // 0.65 & 0.76 // 0.84 & 0.75 // 0.82 \\ \hline
\end{tabular}
\label{fig:tab-statistics}
\end{table}

Further, to avoid occasionality, 5k similar fake image pairs per class are generated by the generator with and without PDPM, and Eq (\ref{eq:revised-latent-sim}) is used for calculating similarity between latent vectors. In Table \ref{fig:tab-statistics}, the value indicates the similarity of latent vector pairs whose corresponding fake images are similar under MSE metrics. It can be seen that PDPM reduces the chance of two different latent vectors mapped to similar fake images.

In GANs, IS~\cite{18,21} and FID~\cite{19} are commonly accepted metrics used for evaluationg the quality and diversity of fake images. On the datasets stated above, 5k fake images per class are generated using the generator with and without PDPM for calculating IS and FID. The parameter $n_{splits}$ of IS is set to 10. Table \ref{tab:score} shows the details. Greater IS value and lower FID value are signs of high quality and diversity of generated data.

\begin{table}[htpb]
\setlength{\abovecaptionskip}{0.2cm}
\begin{spacing}{1.2}
\caption{\textbf{Quantitative results of IS and FID.}}
\label{tab:score}
\begin{tabular}{l|l|lll}
\hline
\multicolumn{2}{l|}{\small{Dataset}}         & \small{WGAN\_GP} & \small{PDPM $\lambda$=5} & \small{PDPM $\lambda$=10} \\ \hline
\multirow{2}{*}{\small{M}}         & $\uparrow$IS  & 2.18$\pm$.003  & 2.19$\pm$.005 & \textbf{2.31$\pm$.005}       \\
                               & $\downarrow$FID & 7.36$\pm$.012      & 6.43$\pm$.009      & \textbf{5.88$\pm$.011}       \\ \hline
\multirow{2}{*}{\small{FM}} & $\uparrow$IS  & 4.28$\pm$.004        & \textbf{4.38$\pm$.006}       & 4.36$\pm$.005       \\
                     & $\downarrow$FID & \textbf{15.68$\pm$.007}  & 15.97$\pm$.013   & 15.72$\pm$.011      \\ \hline
\multirow{2}{*}{\small{C10}}      & $\uparrow$IS  & 7.35$\pm$.007      & 7.52$\pm$.005       & \textbf{7.83$\pm$.007}       \\
                               & $\downarrow$FID & 29.84$\pm$.017       & \textbf{28.45$\pm$.015} & 29.03$\pm$.013 \\ \hline
\multirow{2}{*}{\small{CelebA}}      & $\uparrow$IS  & 2.78$\pm$.002      & 2.91$\pm$.005       & \textbf{2.94$\pm$.002}       \\
                               & $\downarrow$FID & 33.48$\pm$.002       & 25.45$\pm$.015 &  \textbf{24.86$\pm$.002} \\ \hline
\end{tabular}
\end{spacing}
\end{table}

\subsubsection{Evaluation on 2D Synthetic Datasets}
On synthetic dataset, the quantitative evaluation results of mode collapse can be got accurately, because the distribution of data and its modes are known. As prior works, GANs with and without PDPM are evaluated on \textbf{2D Ring} and \textbf{2D Grid}. 2D Ring dataset contains eight 2D Gaussian distributions whose centers locate on a ring equally. 2D Grid contains twenty-five 2D Gaussian distributions whose centers locate on the meshgrid of a square. For comparison, PDPM is applied into the vanilla GAN, Unrolled GAN and BourGAN. The number of modes captured by generator and the percentage of points generated by generator in high-quality (h-q) are used as metrics. As in \cite{26}, a sample is counted as high quality, if it is within three standard deviations of the nearest mode, and the number of modes captured by generator is the number of Gaussian centers which are nearest to at least one high quality sample.

\begin{table}[htpb]
\begin{spacing}{1.15}
\setlength{\abovecaptionskip}{0.2cm}
\caption{\textbf{Quantitative results on 2D Synthetic Dataset.}}
\label{tab:synthetic_table}
\begin{tabular}{lllll}
\hline
                 & \multicolumn{2}{l}{2D Ring} & \multicolumn{2}{l}{2D Grid} \\
                 & modes        & h-q          & modes        & h-q          \\ \hline
GAN              & 1.0          & $\times$        & 17.7         & 82.3         \\
ALI        & 2.8        & 0.13            & 12.8        & 1.6          \\
Unrolled GAN     & 7.6          & 87.97        & 14.9         & 4.89         \\
VEEGAN          & 8.0          & 86.77        & 24.4         & 77.16        \\
PacGAN           & 7.8          & 98.21        & 24.3         & 79.46        \\
BourGAN          & 8.0          & 99.76        & 25.0         & 95.91        \\ \bottomrule
GAN\_PDPM          & 2.0          & $\times$        & 21.3         & 80.8         \\
Unrolled\_PDPM & 8.0          & 99.36        & 21.7         & 75.21        \\
BourGAN\_PDPM      & 8.0          & 99.89        & 25.0         & 95.99        \\ \bottomrule
\end{tabular}
\end{spacing}
\end{table}

\begin{figure*}[htpb]
\begin{center}
   \includegraphics[width=.98\linewidth]{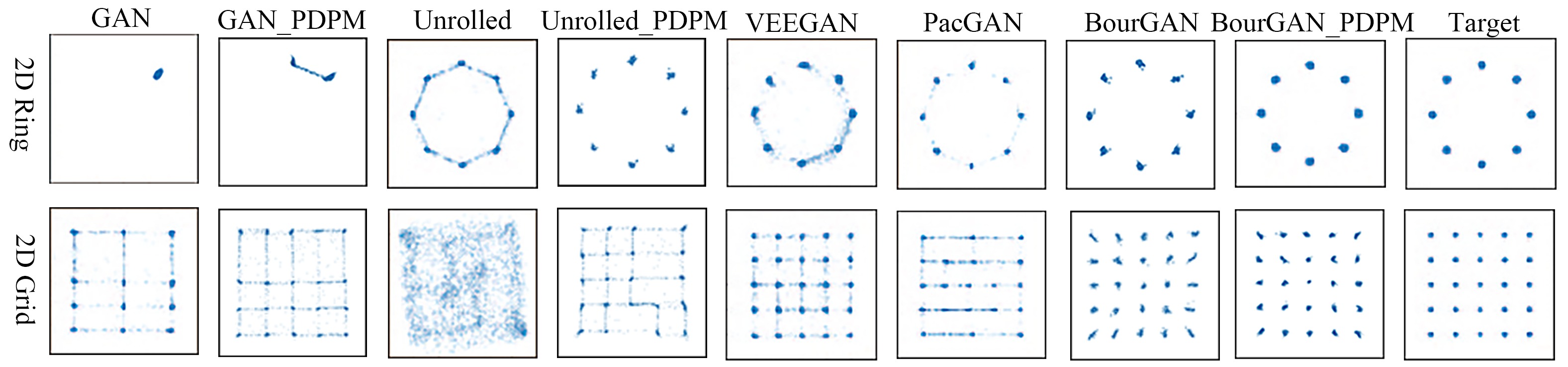}
\end{center}
   \caption{\textbf{Visual results on Synthetic Dataset.} From the first two columns, it can be seen that PDPM help the vanilla GAN capture more modes, especially in 2D Grid, the GAN with PDPM captures four more modes than its vanilla counterpart.}
\label{fig:synthetic}
\end{figure*}

From visual results in Figure~\ref{fig:synthetic} and quantitative results in Table~\ref{tab:synthetic_table}, it can be seen that GAN with PDPM captures more modes of the data distribution, and the vanilla GAN group with PDPM outperforms the ALI and Unrolled GAN on 2D Grid Dataset while closer to VEEGAN and PacGAN. Besides, from BourGAN and BourGAN\_PDPM in Figure \ref{fig:synthetic}, it is clear to see that the group with PDPM converges to the mode centers better than its vanilla counterpart. The results of Unrolled GAN, VEEGAN and PacGAN are from \cite{29}, no official codes of VEEGAN and PacGAN are found until we finish this part, thus, PDPM is not applied into these GANs.

\subsection{Multi Task Applications}
In this part, PDPM is applied into image data augmentation, image generation and domain translation, results in these tasks all suggest that GANs with PDPM outperform their vanilla counterpart.

\subsubsection{Image Generation on CelebA}
The GANs are split into two groups which are DCGAN series with \{DCGAN, DCGAN\_MS, DCGAN\_PDPM\} and WGAN\_GP series with \{WGAN\_GP, WGAN\_GP\_MS, WGAN\_GP\_PDPM\}. Here MS represents the mode seeking regularization propsed in MSGAN. The coefficient $\lambda_{ms}$ is set to 1, and the penalty coefficient $\lambda$ of PDPM shown in Eq (\ref{eq:total-loss}) is set to 10. $\mathbf{Adam}$ with $\beta_1$=0.5 and $\beta_2$=0.9 is used as optimizer, and the learning rate is set to 1e-4. All GANs are trained with a batch size of 128 and 100 epochs in total. The details of architectures are attached in Appendix Table \ref{gans-celeba}. Figure \ref{fig:men-women} shows the results of linear interpolation in latent space, and Table \ref{is-fid-celeba} gives the quantitative results of IS and FID. In Figure \ref{fig:men-women}, MS group generates many noisy pixels, and the transition between images is not smooth since the man with glasses only appears in last two images. By contrast, PDPM can interpolate between two images without noisy pixels, and the transition is much more smoother.

\begin{figure}[htpb]
\begin{center}
    \includegraphics[width=0.45\textwidth]{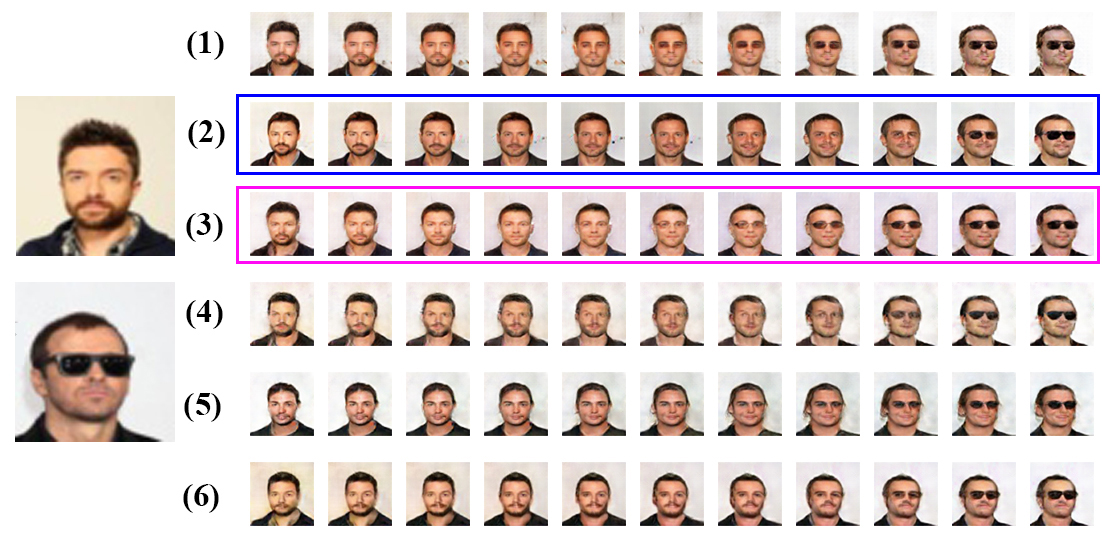}
\end{center}
    \caption{\textbf{Linear interpolation in latent space.} From (1) to (6) are WGAN\_GP, WGAN\_GP\_MS, WGAN\_GP\_PDPM, DCGAN, DCGAN\_MS and DCGAN\_PDPM. The MS group in blue box generates some noisy pixels while PDPM group in purple box not.}
    \label{fig:men-women}
\end{figure}

It can be seen that PDPM also gets higher IS value and lower FID value compared with DCGAN, WGAN\_GP and MSGAN from quantitative results shown in Table \ref{is-fid-celeba}.

\begin{table}[htpb]
\setlength{\abovecaptionskip}{0.2cm}
\begin{spacing}{1.2}
\caption{\textbf{IS and FID results on CelebA.}} 
\label{is-fid-celeba}
\begin{tabular}{lllll}
\hline
\multicolumn{2}{l}{} &\small{DCGAN}    &\small{DCGAN\_MS}    &\small{DCGAN\_PDPM}    \\ \cline{3-5} 
\multicolumn{2}{l}{$\uparrow$\small{IS}}      &2.113$\pm$ 0.014  &2.360$\pm$0.006  & \textbf{2.379$\pm$0.013}              \\
\multicolumn{2}{l}{$\downarrow$\small{FID}}     &24.23$\pm$0.150 &23.51$\pm$0.090 & \textbf{21.76$\pm$0.110}             \\ \hline
\multicolumn{2}{l}{}        &\small{WGAN\_GP} & \small{WGAN\_GP\_MS} & \small{WGAN\_GP\_PDPM} \\ \cline{3-5} 
\multicolumn{2}{l}{$\uparrow$\small{IS}}      &2.775$\pm$0.018    & 2.927$\pm$0.016 & \textbf{2.941$\pm$0.021} \\
\multicolumn{2}{l}{$\downarrow$\small{FID}}     &33.48$\pm$0.011 & 24.86$\pm$0.020 & \textbf{24.18$\pm$0.031}\\ \hline
\end{tabular}
\end{spacing}
\end{table}

\subsubsection{Image Data Augmentation with PDPM}
GANs with PDPM are used for augmentating data on MNIST, Fashion-MNIST and CIFAR-10. The fake images are served as auxiliary training set. ResNet20 proposed in \cite{20} is adopted as classification net. $\mathbf{SGD}$ optimizer is used with learning rate decay. 
Results of accuracy on testing set are shown in Table~\ref{acc-table}, training details are attached in Appendix Figure~\ref{fig:resnet-acc}.

\begin{table}[htpb]
\setlength{\abovecaptionskip}{0.2cm}
\begin{spacing}{1.2}
\caption{\textbf{Testing Accuracy on Several Datasets.}}
\label{acc-table}
\begin{tabular}{l|lll}
\hline
Testing Acc       & MNIST & Fashion-MNIST & CIFAR-10 \\ \hline
Baseline          & 0.9897    & 0.9257           & $\times$         \\
DA & $\times$    & $\times$               & 0.9172         \\
WGAN\_GP          & 0.9951     & 0.9394            & 0.9184        \\
WGAN\_GP\_MS          & 0.9961     & 0.9430            & 0.9200        \\
PDPM\_1 $\lambda=5$          & \textbf{0.9975}     &0.9465             & \textbf{0.9239}        \\
PDPM\_2    $\lambda=10$       & 0.9969     & \textbf{0.9527}            & 0.9212        \\ \hline
\end{tabular}
DA : Traditional Data Augmentation
\end{spacing}
\end{table}

Compared with WGAN\_GP, PDPM gains improvements of 0.24\%, 1.33\% and 0.55\% on MNIST, Fashion-MNIST and CIFAR-10 respectively. More details about the training process refer to Appendix Figure~\ref{fig:resnet-acc}.

\section{Conclusion}
In this paper, a pluggable block called diversity penalty module (PDPM) has been proposed to alleviate mode collapse in GAN. This penalty term is used to enforce the similarity between feature pairs to be consistent with that between latent vector pairs. The advantage of our proposed method is its generalization ability, it almost can be combined with all GANs in different architectures and vision tasks. 

\bibliography{aaai22}

\clearpage
\onecolumn
\section{Appendix}

\begin{figure*}[htpb]
	\subfigure[Testing Acc on MNIST]{
        \includegraphics[width=0.325\textwidth]{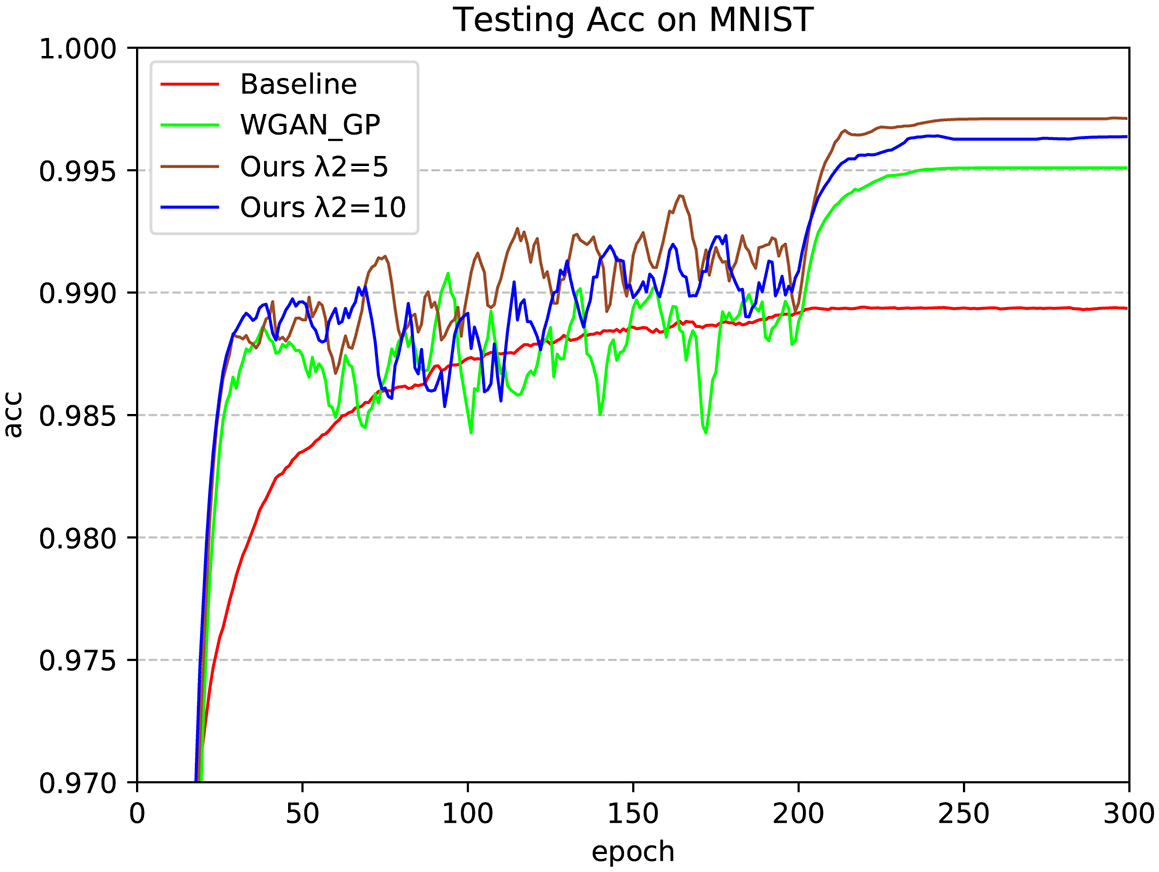}
    }
    \subfigure[Testing Acc on Fashion-MNIST]{
        \includegraphics[width=0.325\textwidth]{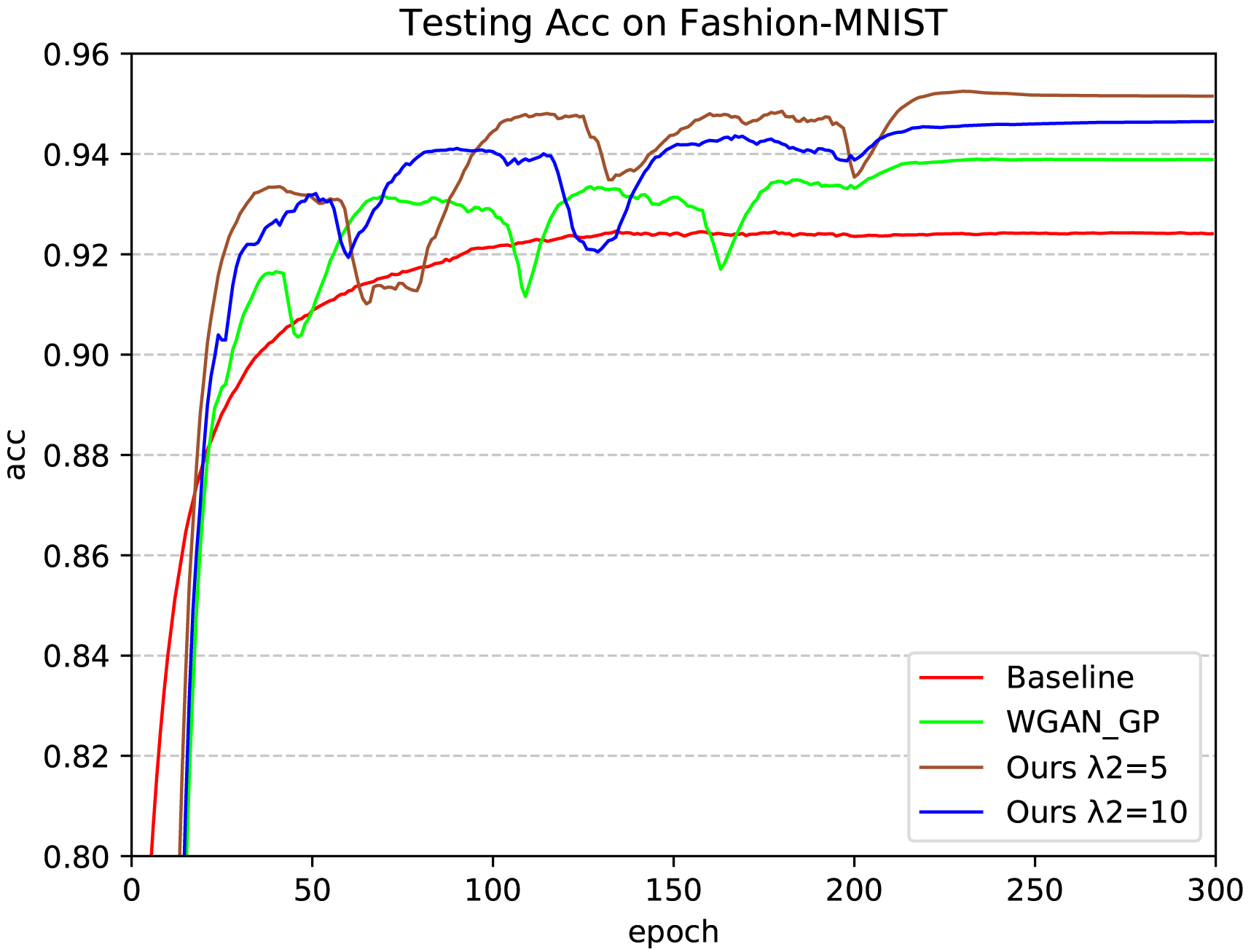}
    }
    \subfigure[Testing Acc on CIFAR-10]{
       \includegraphics[width=0.325\textwidth]{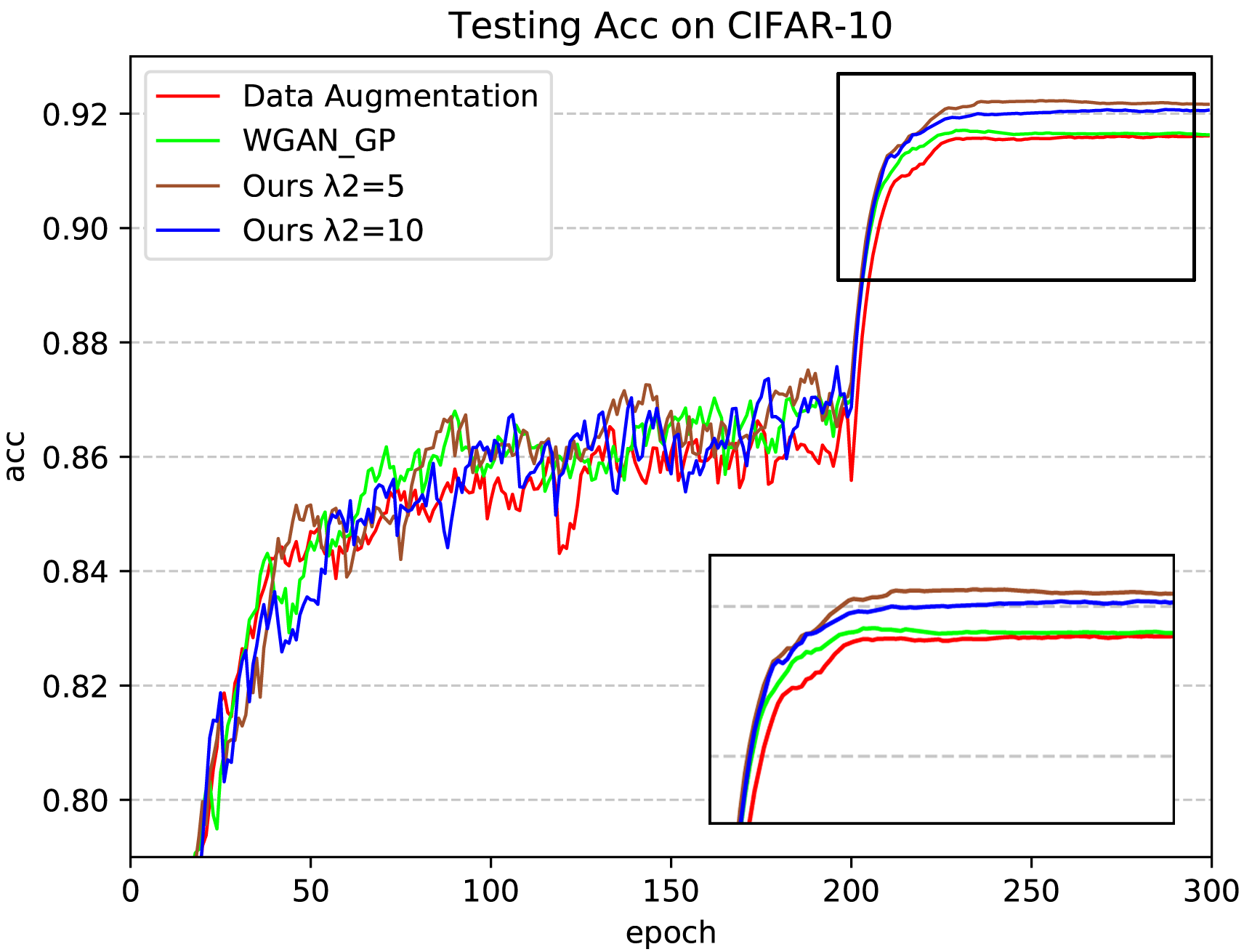}
    }
    \caption{\textbf{Image Classification using GAN-based Data Augmentation on Several Datasets.}}
    \label{fig:resnet-acc}
\end{figure*}

\begin{table*}[htpb]
\setlength{\abovecaptionskip}{0.2cm}
\begin{spacing}{1.3}
\begin{center}
\caption{\textbf{Architectures of GAN in Keras-like style on MNIST / Fashion-MNIST / CIFAR-10.}} 
\label{architecture}
\begin{tabular}{|l|l|l|l|l|l|l|}
\hline
\multicolumn{3}{|l|}{Names}                                              & \multicolumn{2}{l|}{MNIST / Fashion-MNIST}                                                    & \multicolumn{2}{l|}{CIFAR-10}                                                              \\ \hline
\multirow{9}{*}{Generator}     & \multirow{2}{*}{Layer\_1} & Input Size  & \multirow{2}{*}{\begin{tabular}[c]{@{}l@{}}Dense\\ BN ReLU Reshape\end{tabular}}   & 100x1    & \multirow{2}{*}{\begin{tabular}[c]{@{}l@{}}Dense\\ ReLU Reshape\end{tabular}}   & 128x1    \\ \cline{3-3} \cline{5-5} \cline{7-7} 
                               &                           & Output Size &                                                                                    & 7x7x24   &                                                                                 & 4x4x512 \\ \cline{2-7} 
                               & \multirow{2}{*}{Layer\_2} & Input Size  & \multirow{2}{*}{\begin{tabular}[c]{@{}l@{}}Conv2DTranspose\\ BN ReLU\end{tabular}} & 7x7x24   & \multirow{2}{*}{\begin{tabular}[c]{@{}l@{}}Conv2DTranspose\\ ReLU\end{tabular}} & 4x4x512  \\ \cline{3-3} \cline{5-5} \cline{7-7} 
                               &                           & Output Size &                                                                                    & 14x14x12 &                                                                                 & 8x8x256   \\ \cline{2-7} 
                               & \multirow{2}{*}{Layer\_3} & Input Size  & \multirow{2}{*}{\begin{tabular}[c]{@{}l@{}}Conv2DTranspose\\ BN ReLU\end{tabular}} & 14x14x12 & \multirow{2}{*}{\begin{tabular}[c]{@{}l@{}}Conv2DTranspose\\ ReLU\end{tabular}} & 8x8x256   \\ \cline{3-3} \cline{5-5} \cline{7-7} 
                               &                           & Output Size &                                                                                    & 14x14x6  &                                                                                 & 16x16x128 \\ \cline{2-7} 
                               & \multirow{2}{*}{Layer\_4} & Input Size  & \multirow{2}{*}{\begin{tabular}[c]{@{}l@{}}Conv2DTranspose\\ Tanh\end{tabular}}    & 14x14x6  & \multirow{2}{*}{\begin{tabular}[c]{@{}l@{}}Conv2DTranspose\\ Tanh\end{tabular}} & 16x16x128 \\ \cline{3-3} \cline{5-5} \cline{7-7} 
                               &                           & Output Size &                                                                                    & 28x28x1  &                                                                                 & 32x32x3  \\ \cline{2-7} 
                               & \multicolumn{2}{l|}{Trainable Params}   & \multicolumn{2}{l|}{124477}                                                                   & \multicolumn{2}{l|}{5162755}                                                                \\ \hline
\multicolumn{7}{|l|}{}                                                                                                                                                                                                                                                \\ \hline
\multirow{9}{*}{Discriminator} & \multirow{2}{*}{Layer\_1} & Input Size  & \multirow{2}{*}{\begin{tabular}[c]{@{}l@{}}Conv2D\\ LeakyReLU\end{tabular}}        & 28x28x1  & \multirow{2}{*}{\begin{tabular}[c]{@{}l@{}}Conv2D\\ LeakyReLU\end{tabular}}     & 32x32x3  \\ \cline{3-3} \cline{5-5} \cline{7-7} 
                               &                           & Output Size &                                                                                    & 14x14x6  &                                                                                 & 16x16x128 \\ \cline{2-7} 
                               & \multirow{2}{*}{Layer\_2} & Input Size  & \multirow{2}{*}{\begin{tabular}[c]{@{}l@{}}Conv2D\\ BN LeakyReLU\end{tabular}}     & 14x14x6  & \multirow{2}{*}{\begin{tabular}[c]{@{}l@{}}Conv2D\\ LeakyReLU\end{tabular}}     & 16x16x128 \\ \cline{3-3} \cline{5-5} \cline{7-7} 
                               &                           & Output Size &                                                                                    & 7x7x12   &                                                                                 & 8x8x256   \\ \cline{2-7} 
                               & \multirow{2}{*}{Layer\_3} & Input Size  & \multirow{2}{*}{\begin{tabular}[c]{@{}l@{}}Conv2D\\ LeakyReLU\end{tabular}}        & 7x7x12   & \multirow{2}{*}{\begin{tabular}[c]{@{}l@{}}Conv2D\\ LeakyReLU\end{tabular}}     & 8x8x256  \\ \cline{3-3} \cline{5-5} \cline{7-7} 
                               &                           & Output Size &                                                                                    & 4x4x24   &                                                                                 & 4x4x512  \\ \cline{2-7} 
                               & \multirow{2}{*}{Layer\_4} & Input Size  & \multirow{2}{*}{\begin{tabular}[c]{@{}l@{}}Flatten\\ Dense\end{tabular}}           & 4x4x512   & \multirow{2}{*}{\begin{tabular}[c]{@{}l@{}}Flatten\\ Dense\end{tabular}}        & 4x4x512  \\ \cline{3-3} \cline{5-5} \cline{7-7} 
                               &                           & Output Size &                                                                                    & 1        &                                                                                 & 1        \\ \cline{2-7} 
                               & \multicolumn{2}{l|}{Trainable Params}   & \multicolumn{2}{l|}{9649}                                                                     & \multicolumn{2}{l|}{4114689}                                                                \\ \hline
\end{tabular}
\end{center}
\end{spacing}
\end{table*}

\begin{table}[htpb]
\setlength{\abovecaptionskip}{0.2cm}
\begin{spacing}{1.5}
\begin{center}
\caption{\textbf{Architectures of GAN in Keras-like style on CelebA.}} 
\label{gans-celeba}
\begin{tabular}{|l|l|l|l|l|l|l|}
\hline
\multicolumn{3}{|l|}{Names}                                               & \multicolumn{2}{l|}{DCGAN Series}                                             & \multicolumn{2}{l|}{WGAN\_GP Series}                                 \\ \hline
\multirow{11}{*}{Generator}     & \multirow{2}{*}{Layer\_1} & Input Size  & \multirow{2}{*}{\begin{tabular}[c]{@{}l@{}}Dense Reshape\\ BN ReLU\end{tabular}}   & 100x1     & \multirow{2}{*}{\begin{tabular}[c]{@{}l@{}}Dense\\ ReLU Reshape\end{tabular}}   & 100x1     \\ \cline{3-3} \cline{5-5} \cline{7-7} 
                                &                           & Output Size &                                                                                    & 4x4x1024  &                                                                                 & 4x4x1024  \\ \cline{2-7} 
                                & \multirow{2}{*}{Layer\_2} & Input Size  & \multirow{2}{*}{\begin{tabular}[c]{@{}l@{}}Conv2DTranspose\\ BN ReLU\end{tabular}} & 4x4x1024  & \multirow{2}{*}{\begin{tabular}[c]{@{}l@{}}Conv2DTranspose\\ ReLU\end{tabular}} & 4x4x1024  \\ \cline{3-3} \cline{5-5} \cline{7-7} 
                                &                           & Output Size &                                                                                    & 8x8x512   &                                                                                 & 8x8x512   \\ \cline{2-7} 
                                & \multirow{2}{*}{Layer\_3} & Input Size  & \multirow{2}{*}{\begin{tabular}[c]{@{}l@{}}Conv2DTranspose\\ BN ReLU\end{tabular}} & 8x8x512   & \multirow{2}{*}{\begin{tabular}[c]{@{}l@{}}Conv2DTranspose\\ ReLU\end{tabular}} & 8x8x512   \\ \cline{3-3} \cline{5-5} \cline{7-7} 
                                &                           & Output Size &                                                                                    & 16x16x256 &                                                                                 & 16x16x256 \\ \cline{2-7} 
                                & \multirow{2}{*}{Layer\_4} & Input Size  & \multirow{2}{*}{\begin{tabular}[c]{@{}l@{}}Conv2DTranspose\\ BN ReLU\end{tabular}} & 16x16x256 & \multirow{2}{*}{\begin{tabular}[c]{@{}l@{}}Conv2DTranspose\\ ReLU\end{tabular}} & 16x16x256 \\ \cline{3-3} \cline{5-5} \cline{7-7} 
                                &                           & Output Size &                                                                                    & 32x32x128 &                                                                                 & 32x32x128 \\ \cline{2-7} 
                                & \multirow{2}{*}{Layer\_5} & Input Size  & \multirow{2}{*}{\begin{tabular}[c]{@{}l@{}}Conv2DTranspose\\ Tanh\end{tabular}}    & 32x32x128 & \multirow{2}{*}{\begin{tabular}[c]{@{}l@{}}Conv2DTranspose\\ Tanh\end{tabular}} & 32x32x128 \\ \cline{3-3} \cline{5-5} \cline{7-7} 
                                &                           & Output Size &                                                                                    & 64x64x3   &                                                                                 & 64x64x3   \\ \cline{2-7} 
                                & \multicolumn{2}{l|}{Trainable Params}   & \multicolumn{2}{l|}{12679171}                                                                  & \multicolumn{2}{l|}{12675331}                                                               \\ \hline
\multicolumn{7}{|l|}{}                                                                                                                                                                                                                                                   \\ \hline
\multirow{11}{*}{Discriminator} & \multirow{2}{*}{Layer\_1} & Input Size  & \multirow{2}{*}{\begin{tabular}[c]{@{}l@{}}Conv2D\\ LeakyReLU\end{tabular}}        & 64x64x3   & \multirow{2}{*}{\begin{tabular}[c]{@{}l@{}}Conv2D\\ LeakyReLU\end{tabular}}     & 64x64x3   \\ \cline{3-3} \cline{5-5} \cline{7-7} 
                                &                           & Output Size &                                                                                    & 32x32x128 &                                                                                 & 32x32x128 \\ \cline{2-7} 
                                & \multirow{2}{*}{Layer\_2} & Input Size  & \multirow{2}{*}{\begin{tabular}[c]{@{}l@{}}Conv2D BN\\ LeakyReLU\end{tabular}}     & 32x32x128 & \multirow{2}{*}{\begin{tabular}[c]{@{}l@{}}Conv2D\\ LeakyReLU\end{tabular}}     & 32x32x128 \\ \cline{3-3} \cline{5-5} \cline{7-7} 
                                &                           & Output Size &                                                                                    & 16x16x256 &                                                                                 & 16x16x256 \\ \cline{2-7} 
                                & \multirow{2}{*}{Layer\_3} & Input Size  & \multirow{2}{*}{\begin{tabular}[c]{@{}l@{}}Conv2D BN\\ LeakyReLU\end{tabular}}     & 16x16x256 & \multirow{2}{*}{\begin{tabular}[c]{@{}l@{}}Conv2D\\ LeakyReLU\end{tabular}}     & 16x16x256 \\ \cline{3-3} \cline{5-5} \cline{7-7} 
                                &                           & Output Size &                                                                                    & 8x8x512 &                                                                                 & 8x8x512 \\ \cline{2-7} 
                                & \multirow{2}{*}{Layer\_4} & Input Size  & \multirow{2}{*}{\begin{tabular}[c]{@{}l@{}}Conv2D BN\\ LeakyReLU\end{tabular}}     & 8x8x512   & \multirow{2}{*}{\begin{tabular}[c]{@{}l@{}}Conv2D\\ LeakyReLU\end{tabular}}     & 8x8x512   \\ \cline{3-3} \cline{5-5} \cline{7-7} 
                                &                           & Output Size &                                                                                    & 4x4x1024  &                                                                                 & 4x4x1024  \\ \cline{2-7} 
                                & \multirow{2}{*}{Layer\_5} & Input Size  & \multirow{2}{*}{\begin{tabular}[c]{@{}l@{}}Flatten\\ Dense Sigmoid\end{tabular}}   & 4x4x1024  & \multirow{2}{*}{\begin{tabular}[c]{@{}l@{}}Flatten\\ Dense\end{tabular}}        & 4x4x1024  \\ \cline{3-3} \cline{5-5}
                                &                           & Output Size &                                                                                    & 8x8x512   &                                                                                 & 8x8x512   \\ \cline{2-7}  \cline{7-7} 
                                &                           & Output Size &                                                                                    & 1         &                                                                                 & 1         \\ \cline{2-7} 
                                & \multicolumn{2}{l|}{Trainable Params}   & \multicolumn{2}{l|}{11038081}                                                                  & \multicolumn{2}{l|}{11034497}                                                               \\ \hline
\end{tabular}
\end{center}
\end{spacing}
\end{table}

\begin{figure*}[htpb]
	\subfigure[Convergence on MNIST]{
        \includegraphics[width=0.315\textwidth]{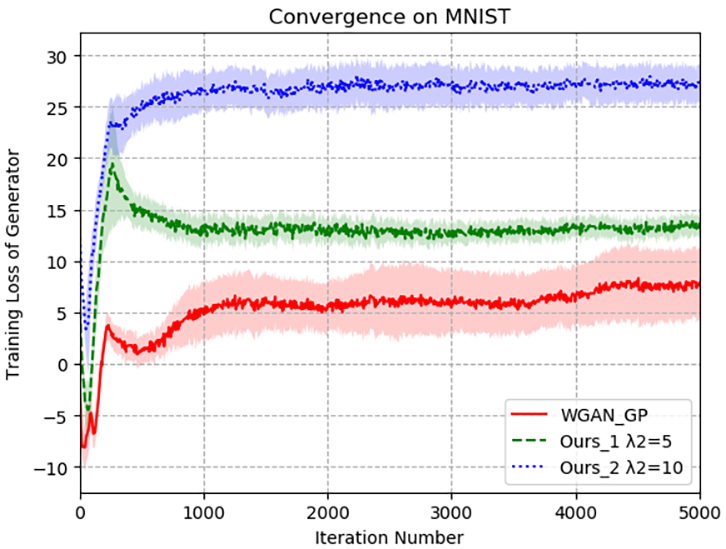}
    }
    \subfigure[Convergence on Fashion-MNIST]{
        \includegraphics[width=0.315\textwidth]{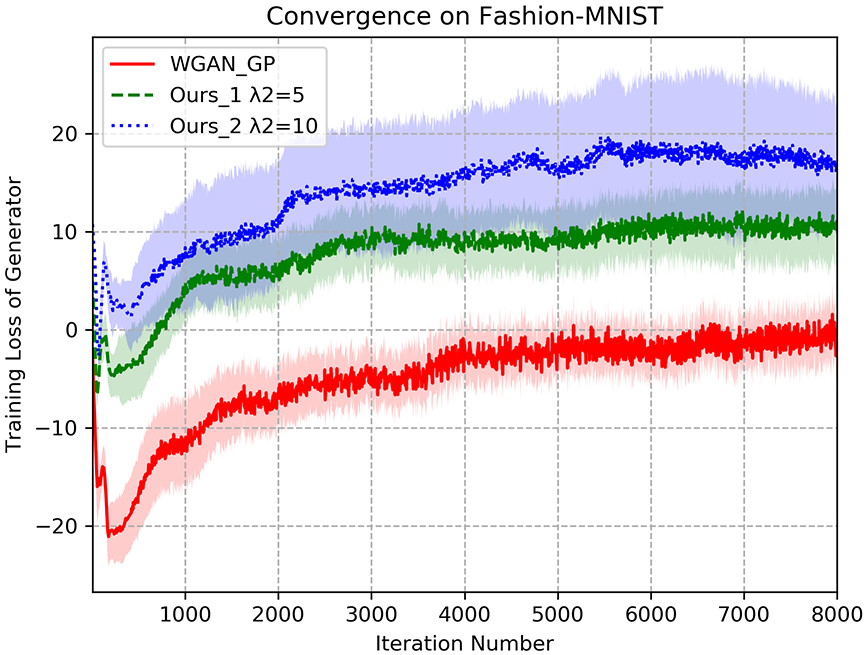}
    }
    \subfigure[Convergence on CIFAR-10]{
       \includegraphics[width=0.32\textwidth]{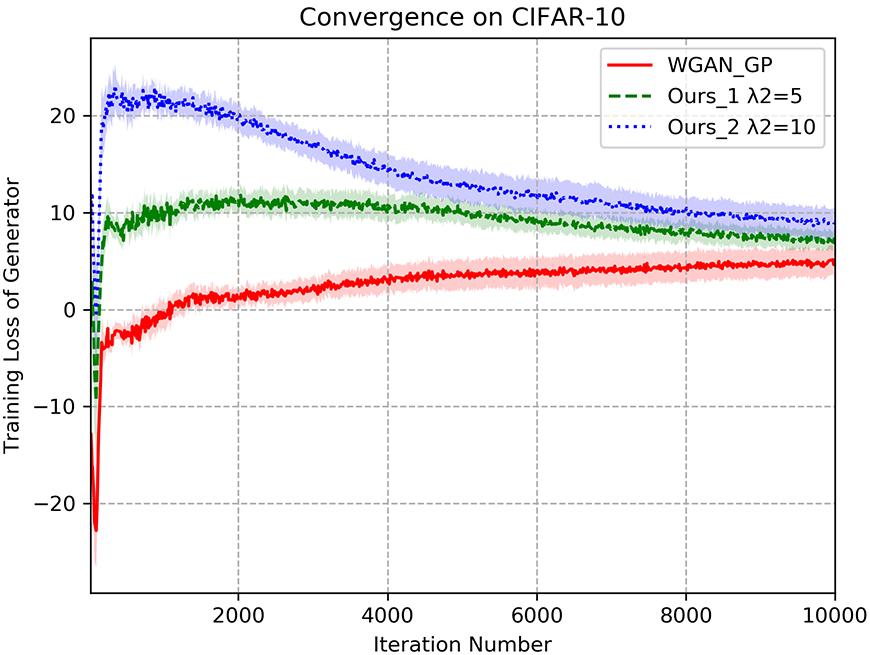}
    }
    \caption{\textbf{Convergence on Several Datasets.} In images above, the lines indicate the averaging loss of 10 generators, the band in light color marks out the boundaries of all losses. Since the loss of discriminator is almost the same in each groups (WGAN\_GP, Ours\_1 and Ours\_2), we just give the results on generator here.}
    \label{fig:convergence}
\end{figure*}

\begin{figure*}[htpb]
	\subfigure[DPM on CelebA]{
        \includegraphics[width=0.4\textwidth]{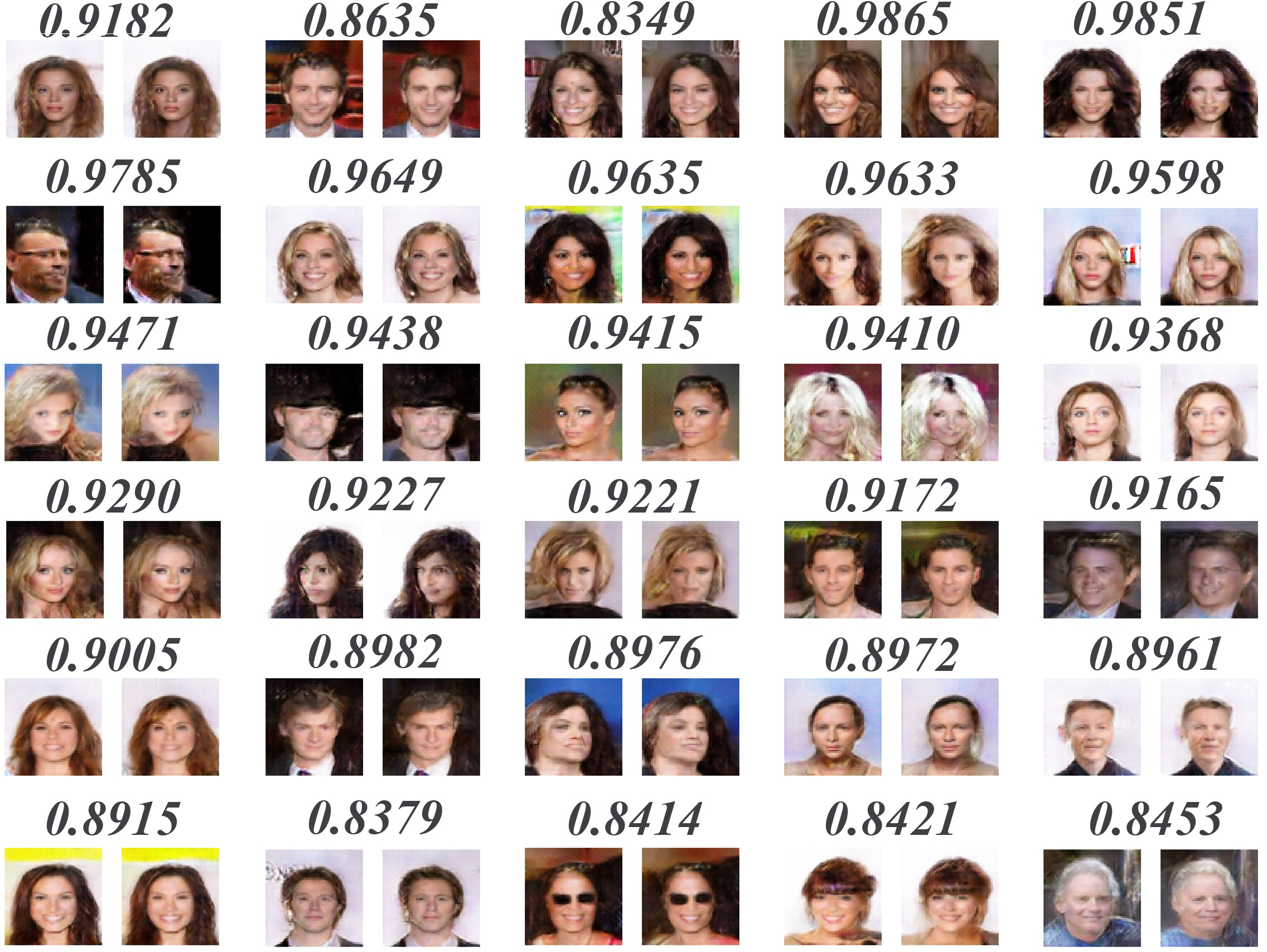}
    }
    \subfigure[DPM on Fashion-MNIST]{
        \includegraphics[width=0.56\textwidth]{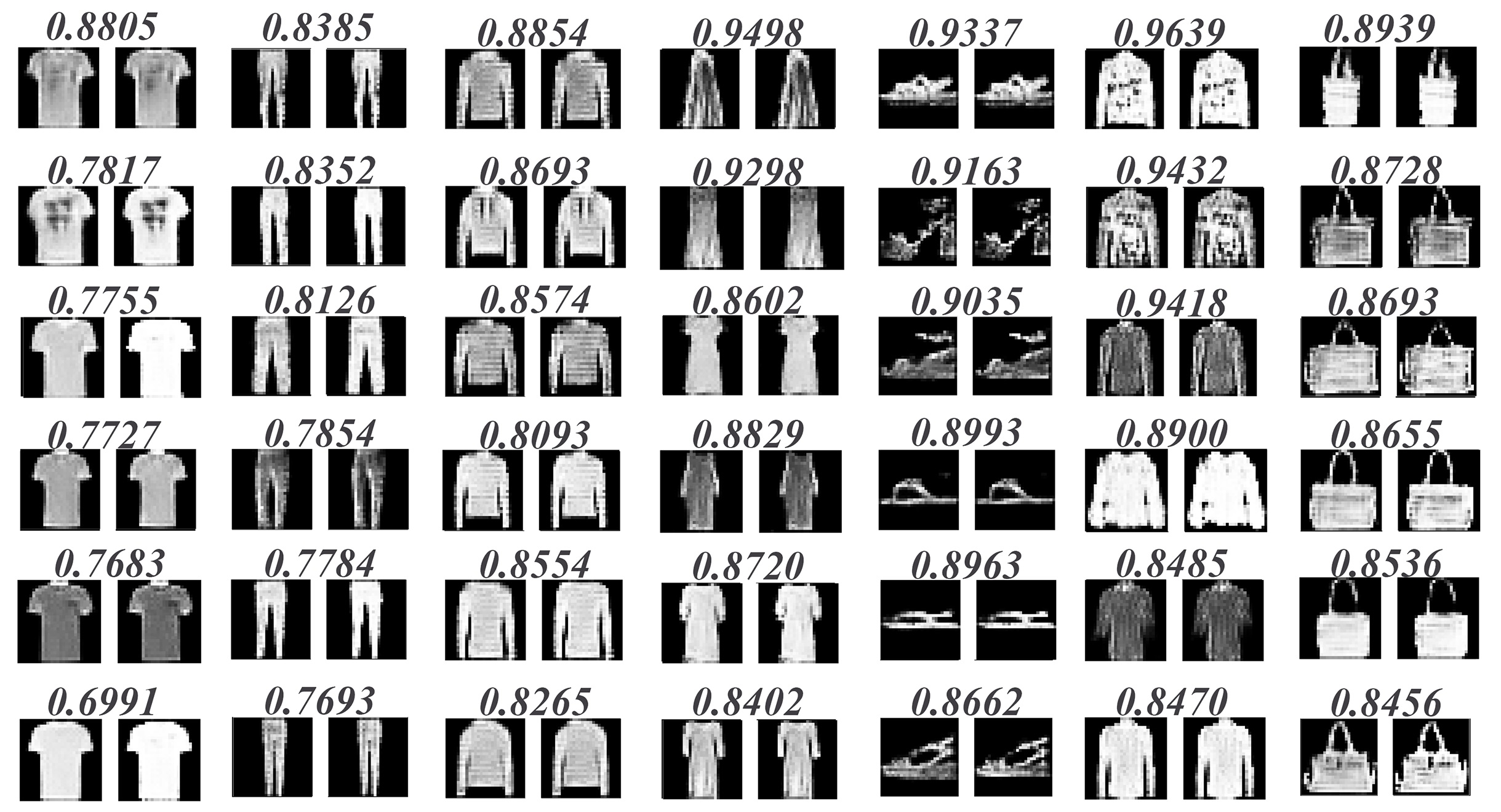}
    }
    \caption{\textbf{Alleviation of mode collapse via DPM on several datasets.}}
    \label{fig:dpm-samples}
\end{figure*}

\begin{figure*}[htpb]
	\subfigure[WGAN\_GP]{
        \includegraphics[width=0.33\textwidth]{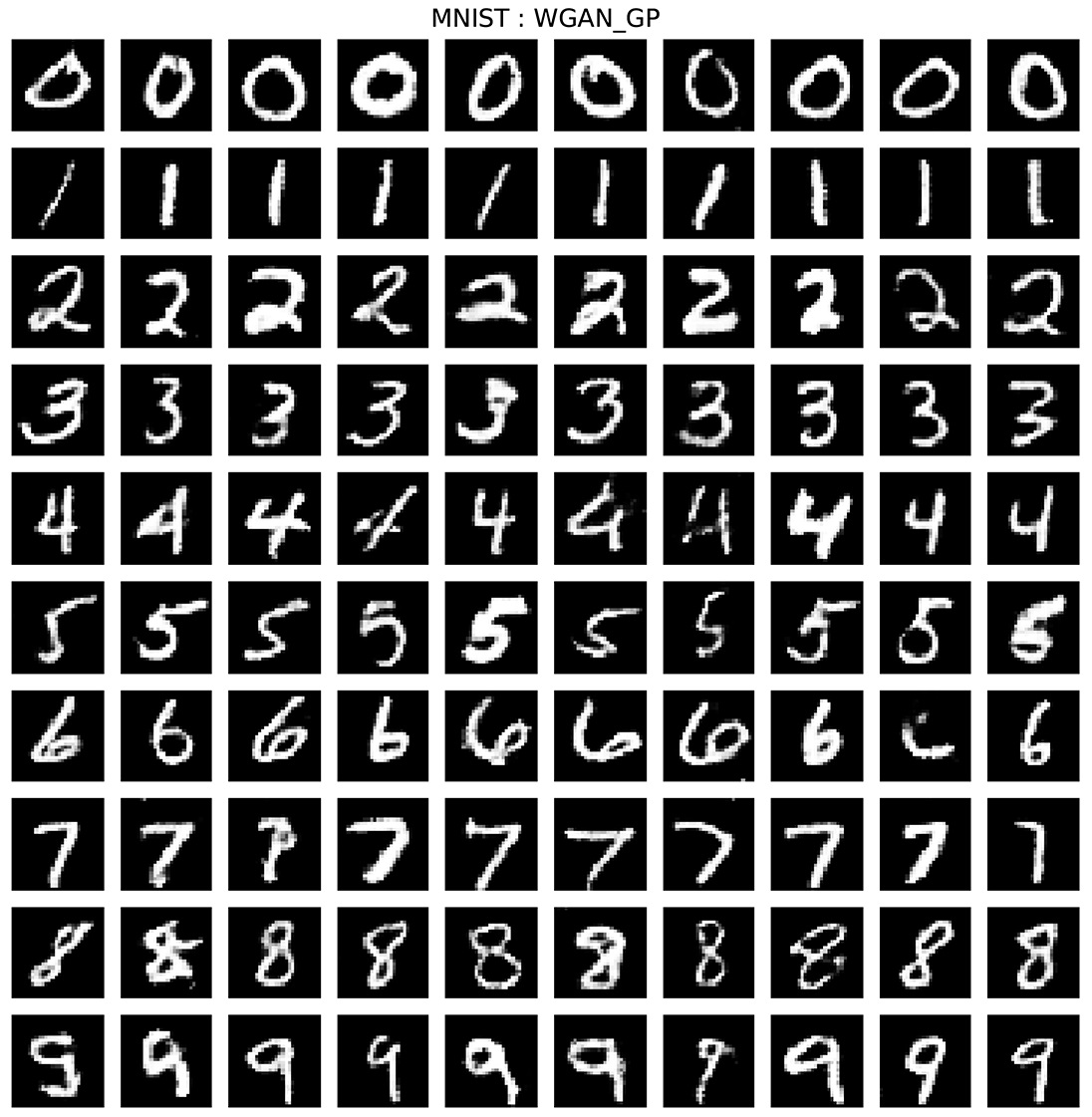}
    }
    \subfigure[WGAN\_GP\_DP with $\lambda_2=5$]{
        \includegraphics[width=0.33\textwidth]{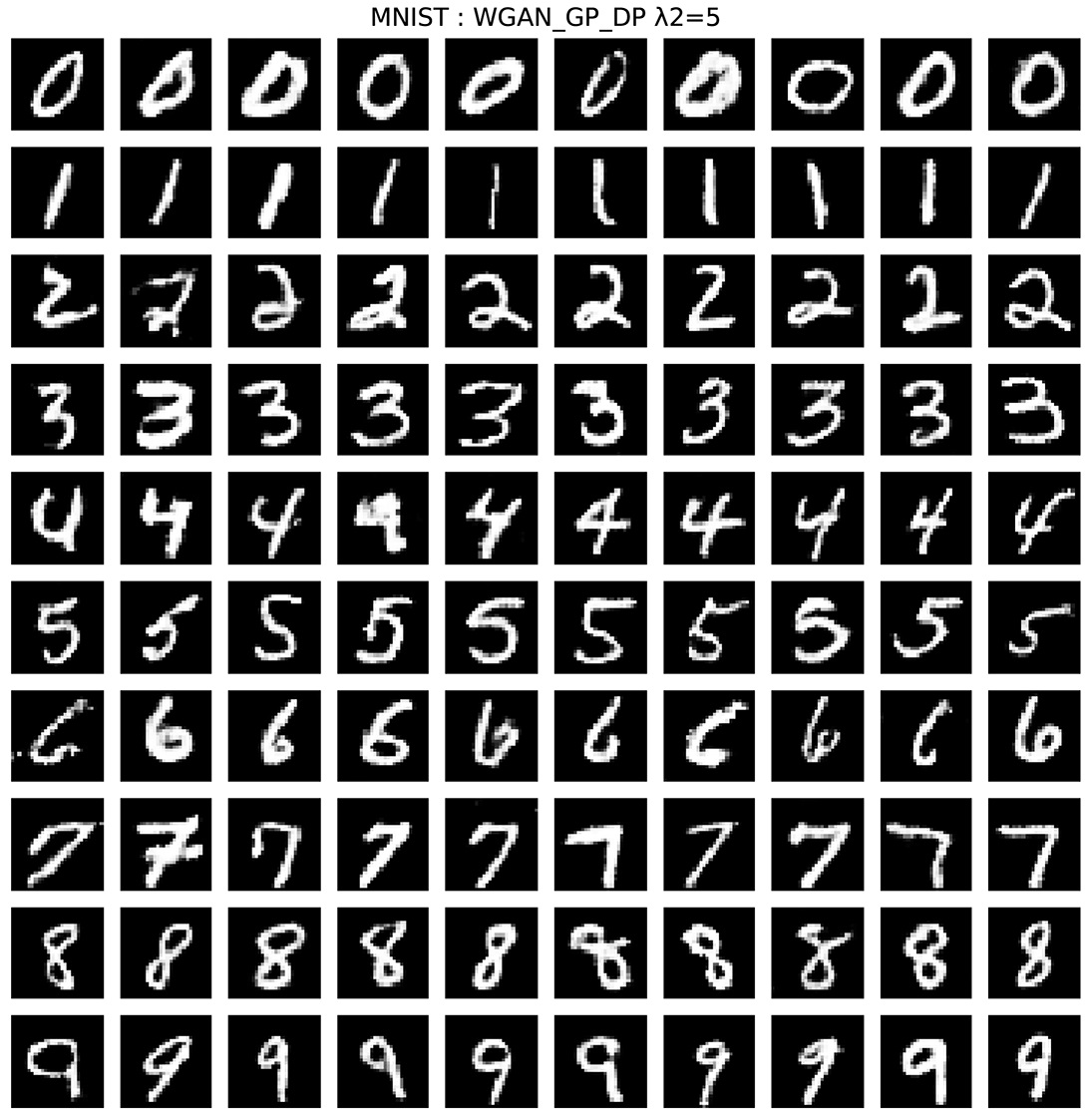}
    }
    \subfigure[WGAN\_GP\_DP with $\lambda_2=10$]{
       \includegraphics[width=0.33\textwidth]{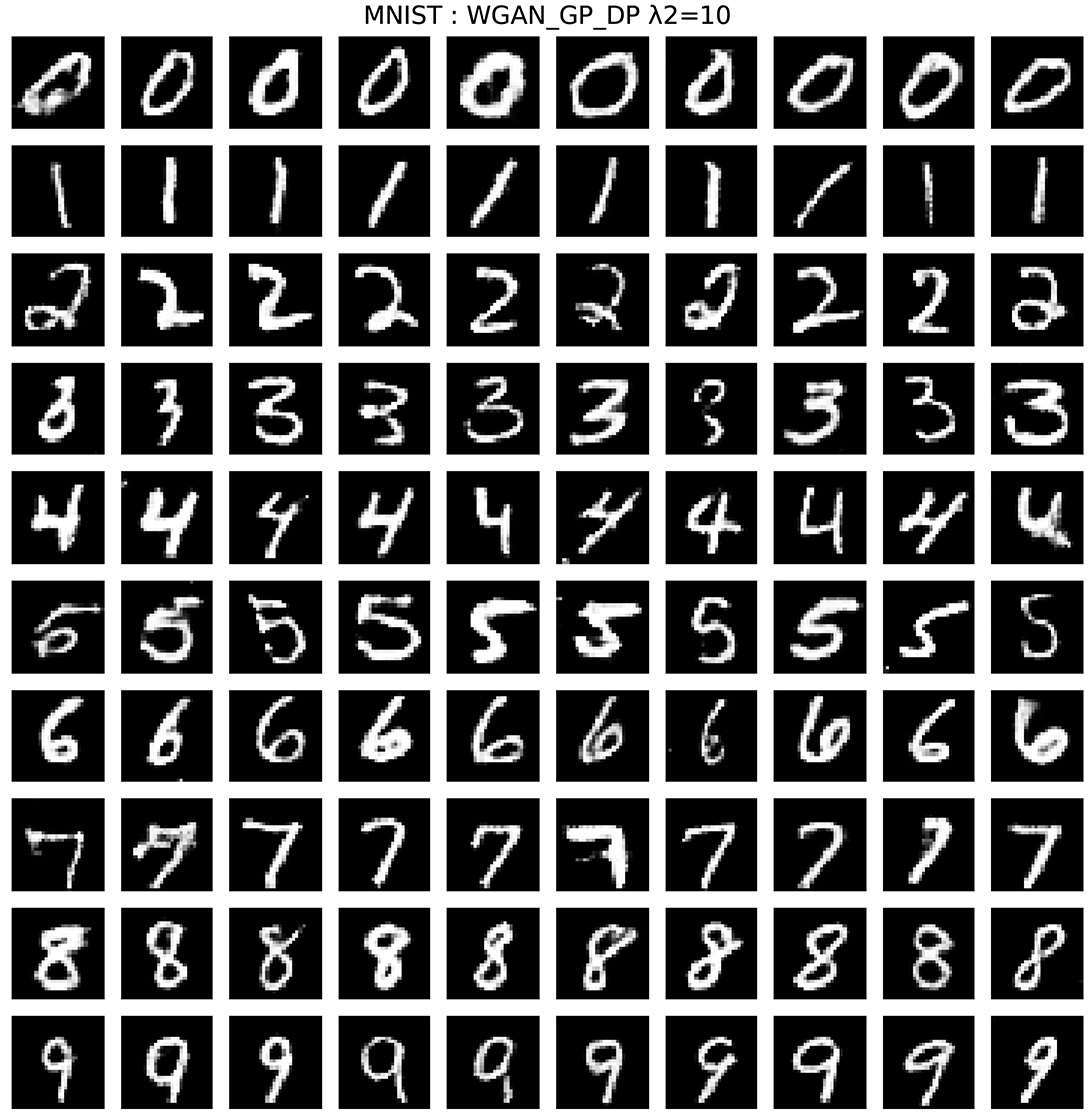}
    }
    \caption{\textbf{Generated Images on MNIST.}}
    \label{fig:mnist_g_imgs}
\end{figure*}

\begin{figure*}[htpb]
	\subfigure[WGAN\_GP]{
        \includegraphics[width=0.33\textwidth]{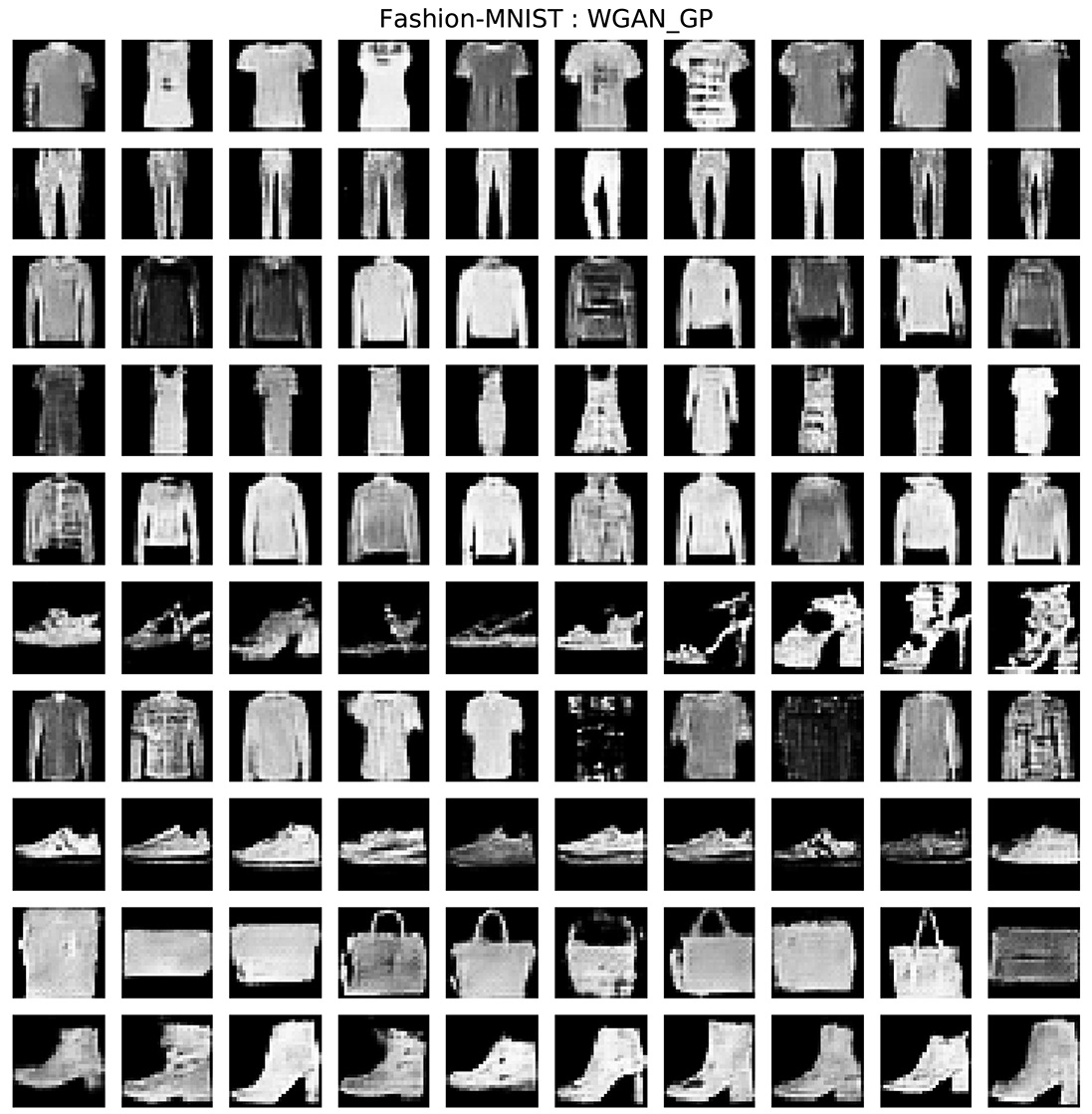}
    }
    \subfigure[WGAN\_GP\_DP with $\lambda_2=5$]{
        \includegraphics[width=0.33\textwidth]{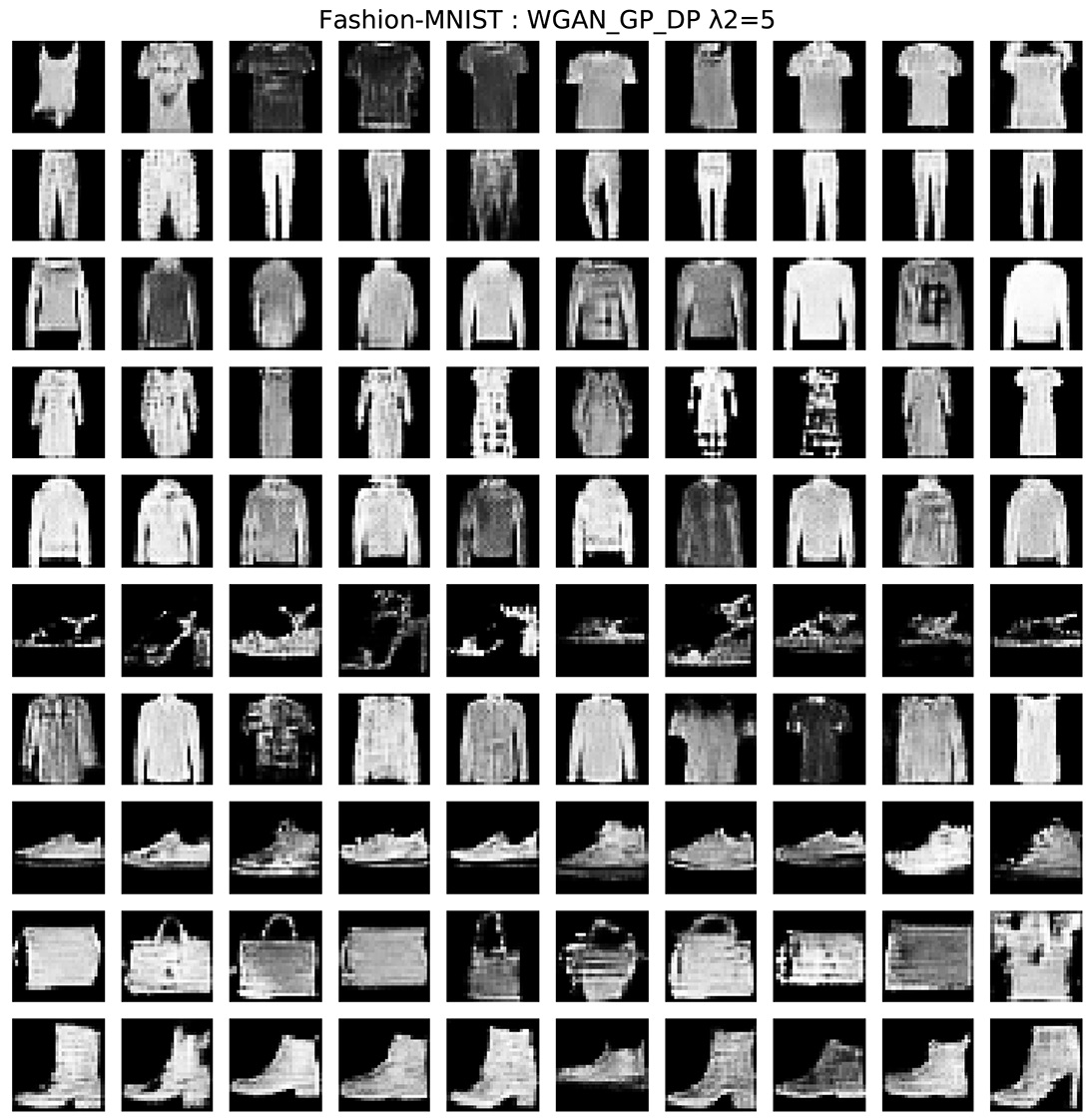}
    }
    \subfigure[WGAN\_GP\_DP with $\lambda_2=10$]{
       \includegraphics[width=0.33\textwidth]{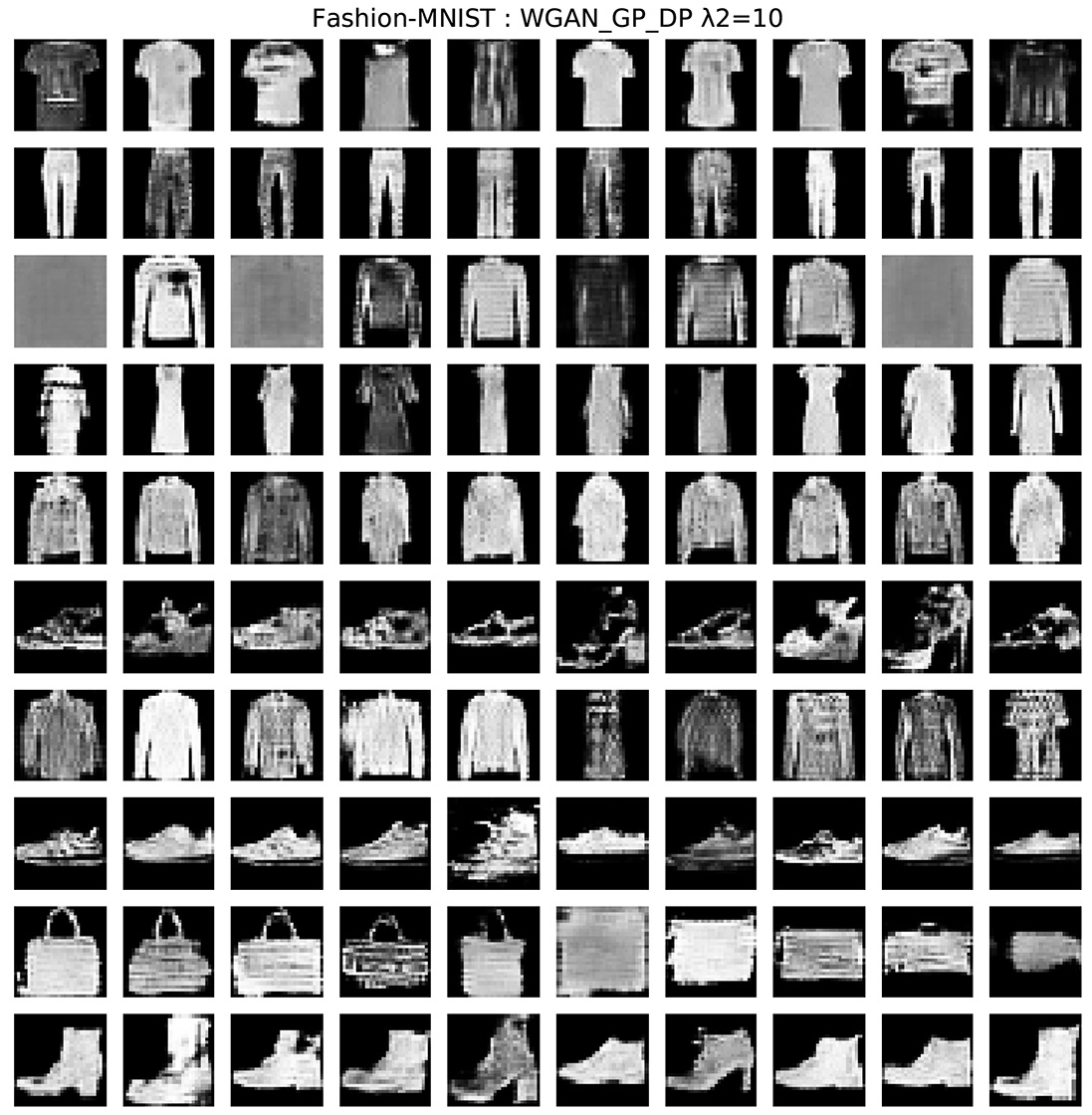}
    }
    \caption{\textbf{Generated Images on Fashion-MNIST.}}
    \label{fig:fm_g_imgs}
\end{figure*}

\begin{figure*}[htpb]
	\subfigure[WGAN\_GP]{
        \includegraphics[width=0.33\textwidth]{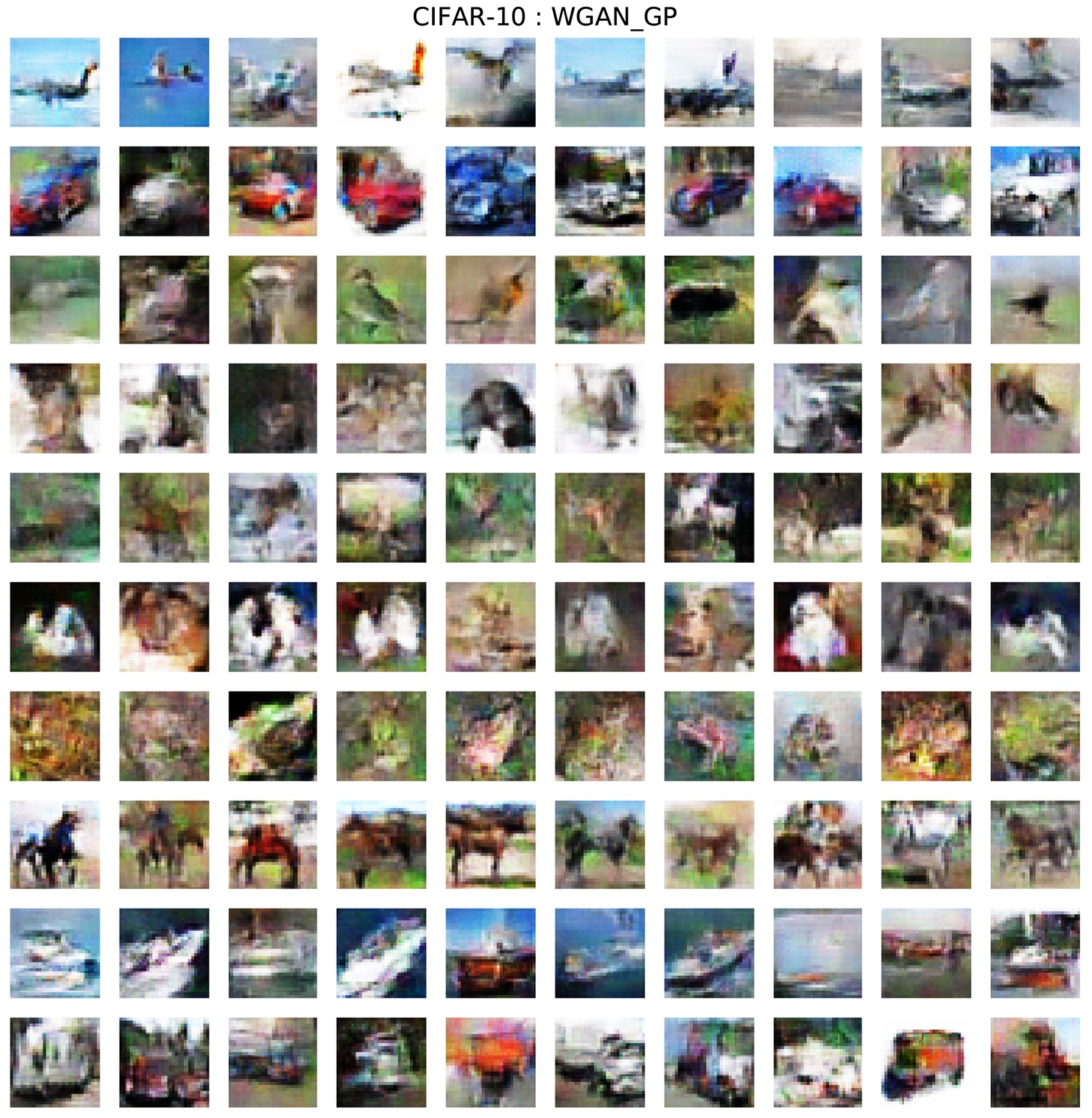}
    }
    \subfigure[WGAN\_GP\_DP with $\lambda_2=5$]{
        \includegraphics[width=0.33\textwidth]{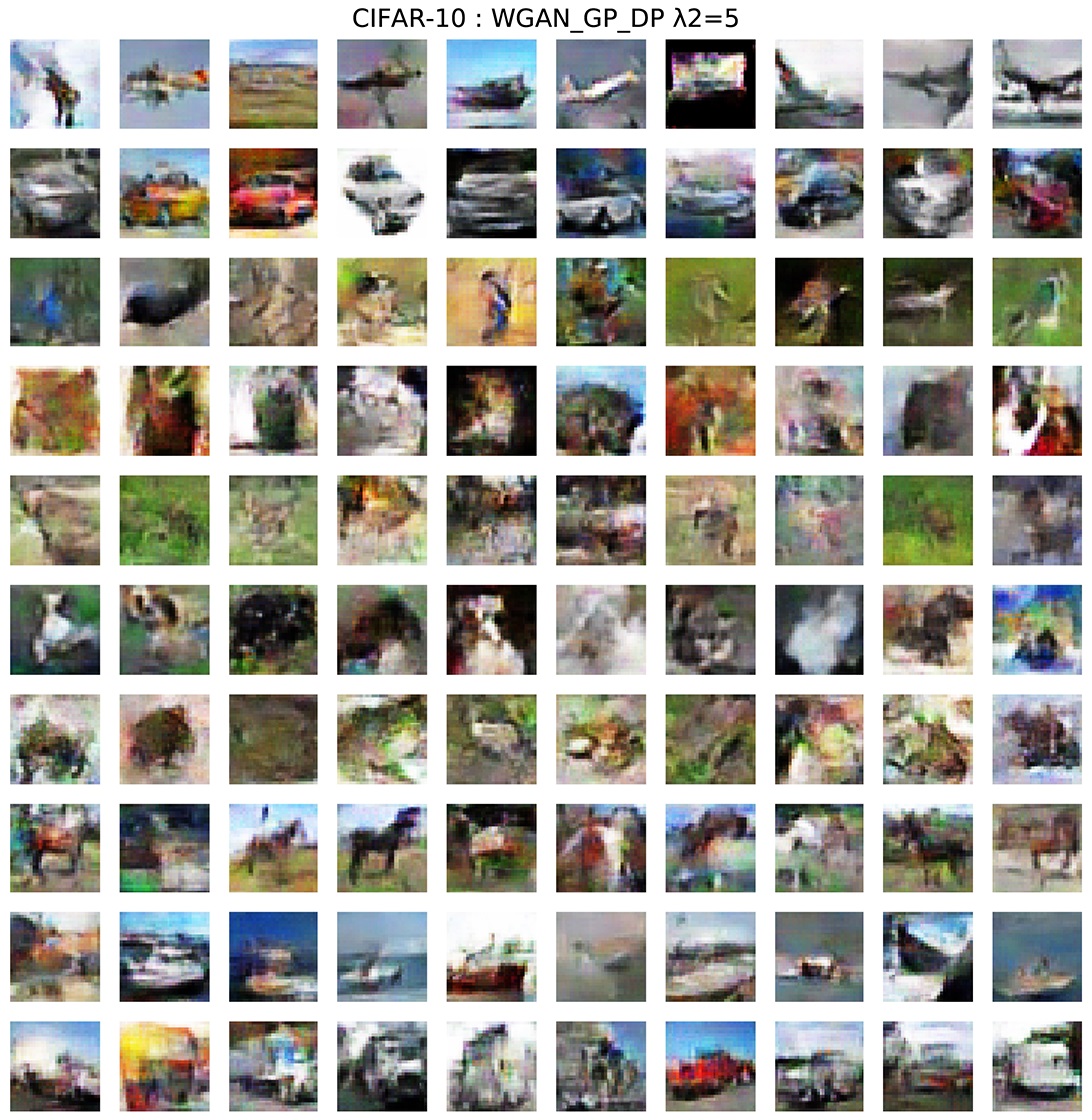}
    }
    \subfigure[WGAN\_GP\_DP with $\lambda_2=10$]{
       \includegraphics[width=0.33\textwidth]{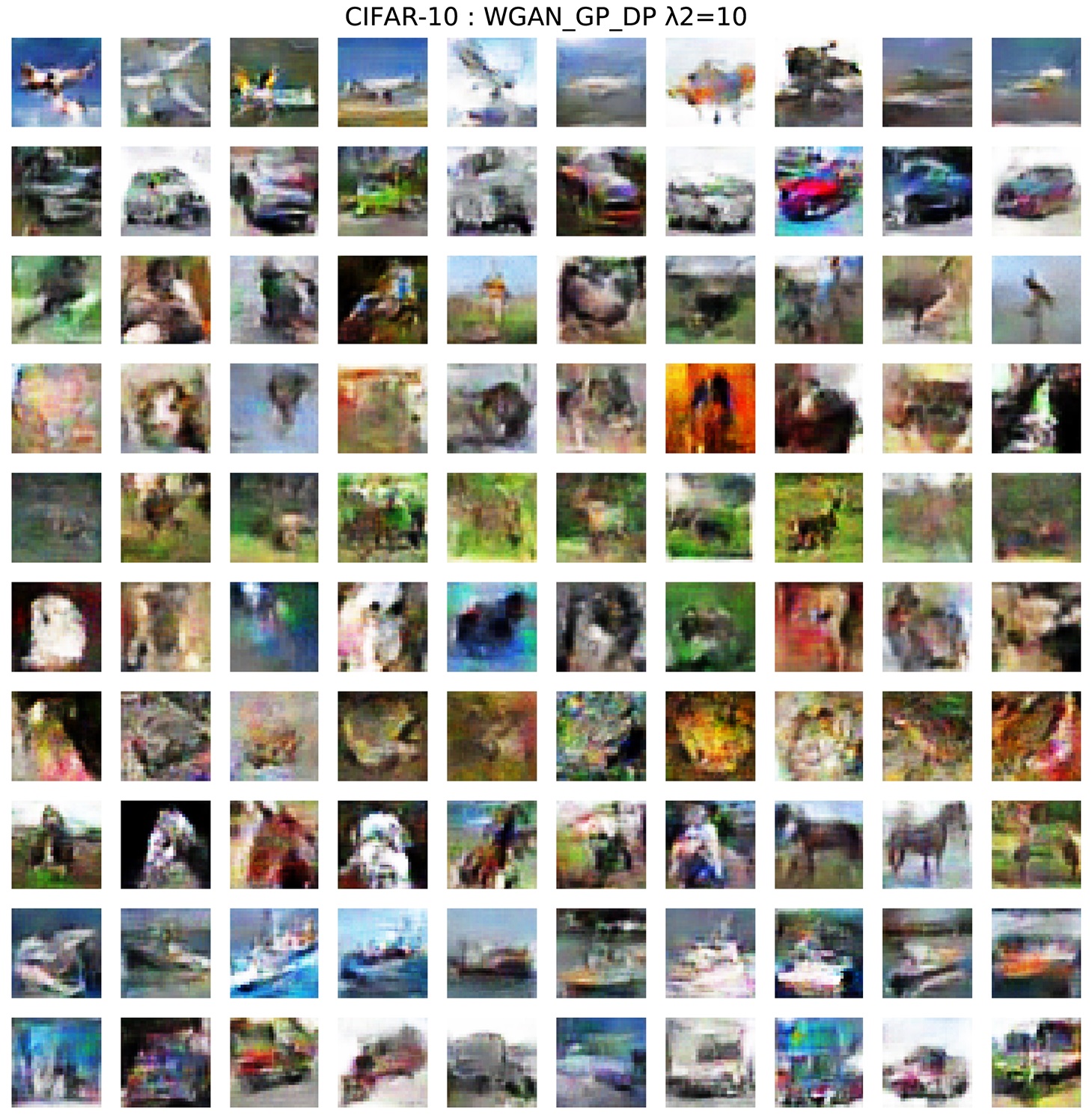}
    }
    \caption{\textbf{Generated Images on CIFAR-10.}}
    \label{fig:cifar_g_imgs}
\end{figure*}

\begin{figure*}[htpb]
	\subfigure[DCGAN]{
        \includegraphics[width=0.33\textwidth]{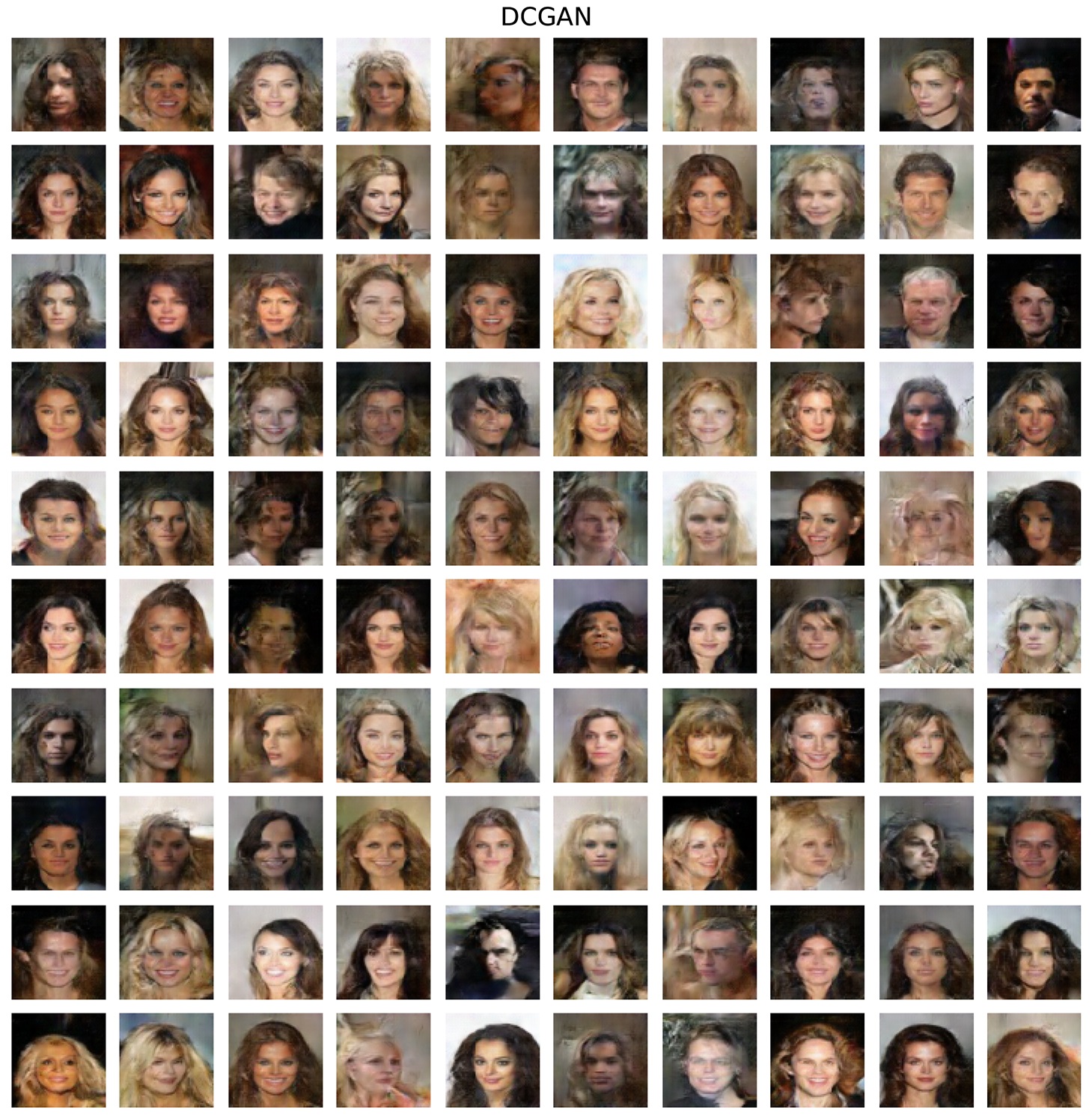}
    }
    \subfigure[DCGAN\_MS]{
        \includegraphics[width=0.33\textwidth]{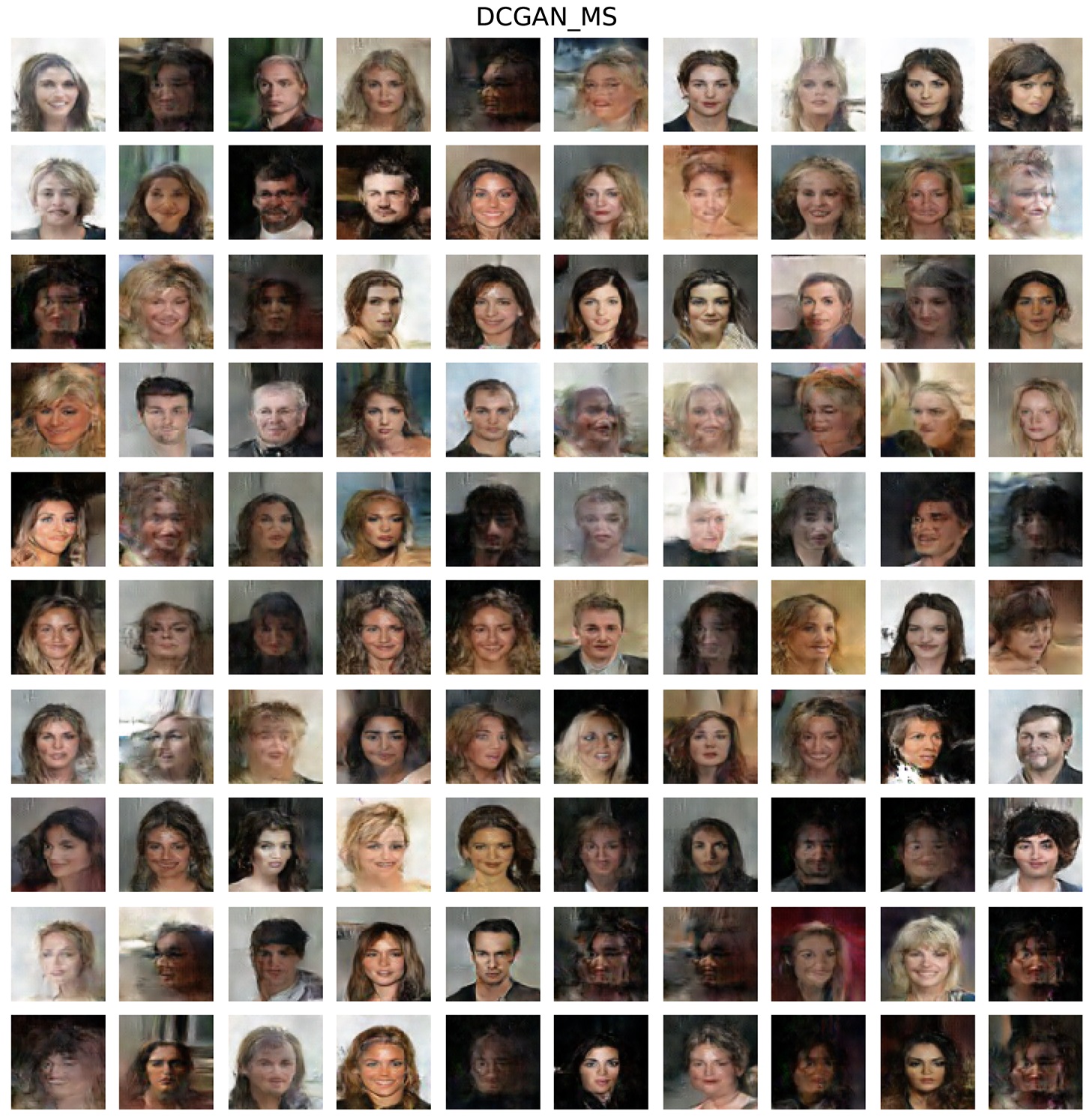}
    }
    \subfigure[DCGAN\_DP with $\lambda_2=10$]{
       \includegraphics[width=0.33\textwidth]{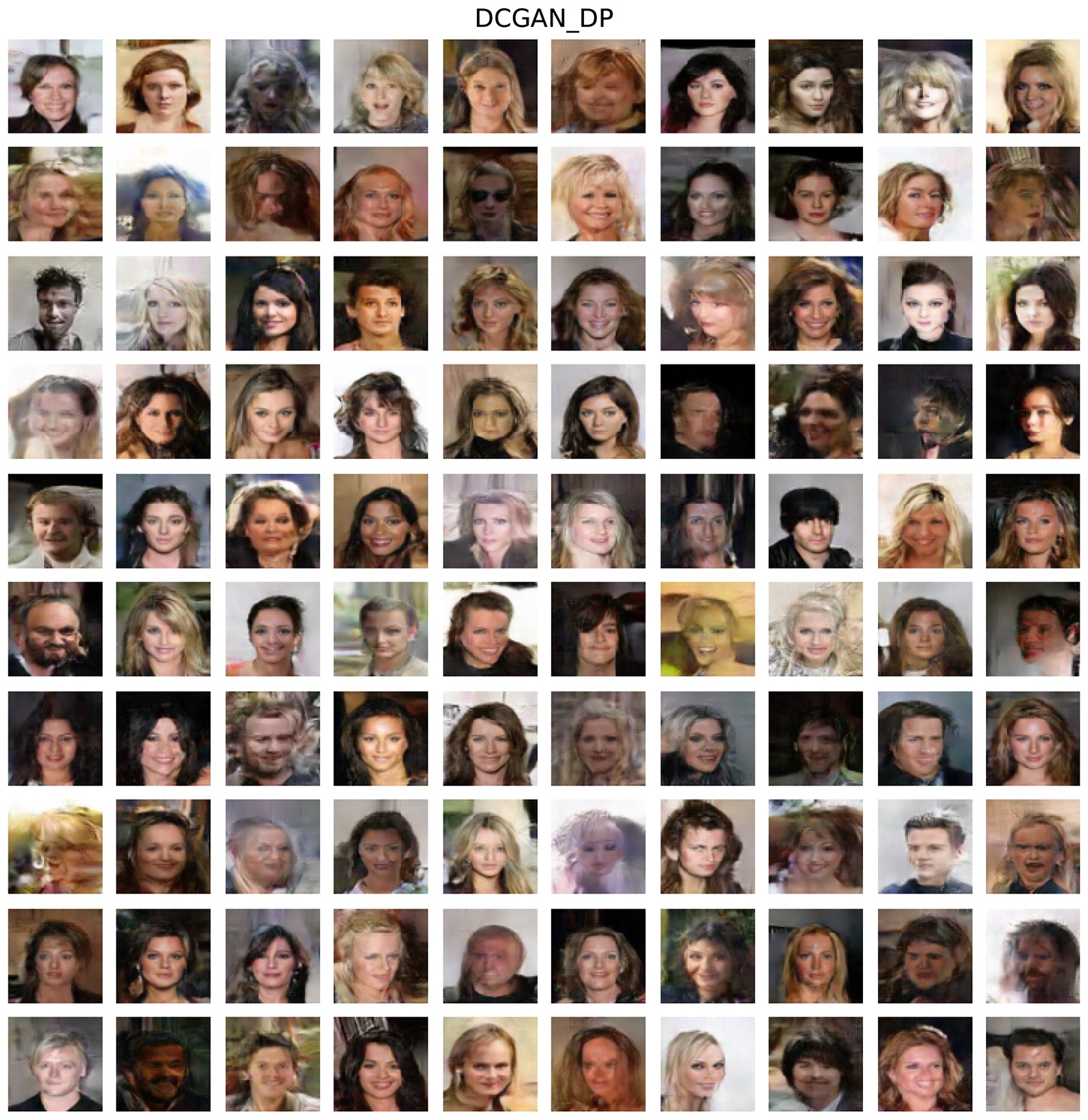}
    }
	\subfigure[WGAN\_GP]{
        \includegraphics[width=0.33\textwidth]{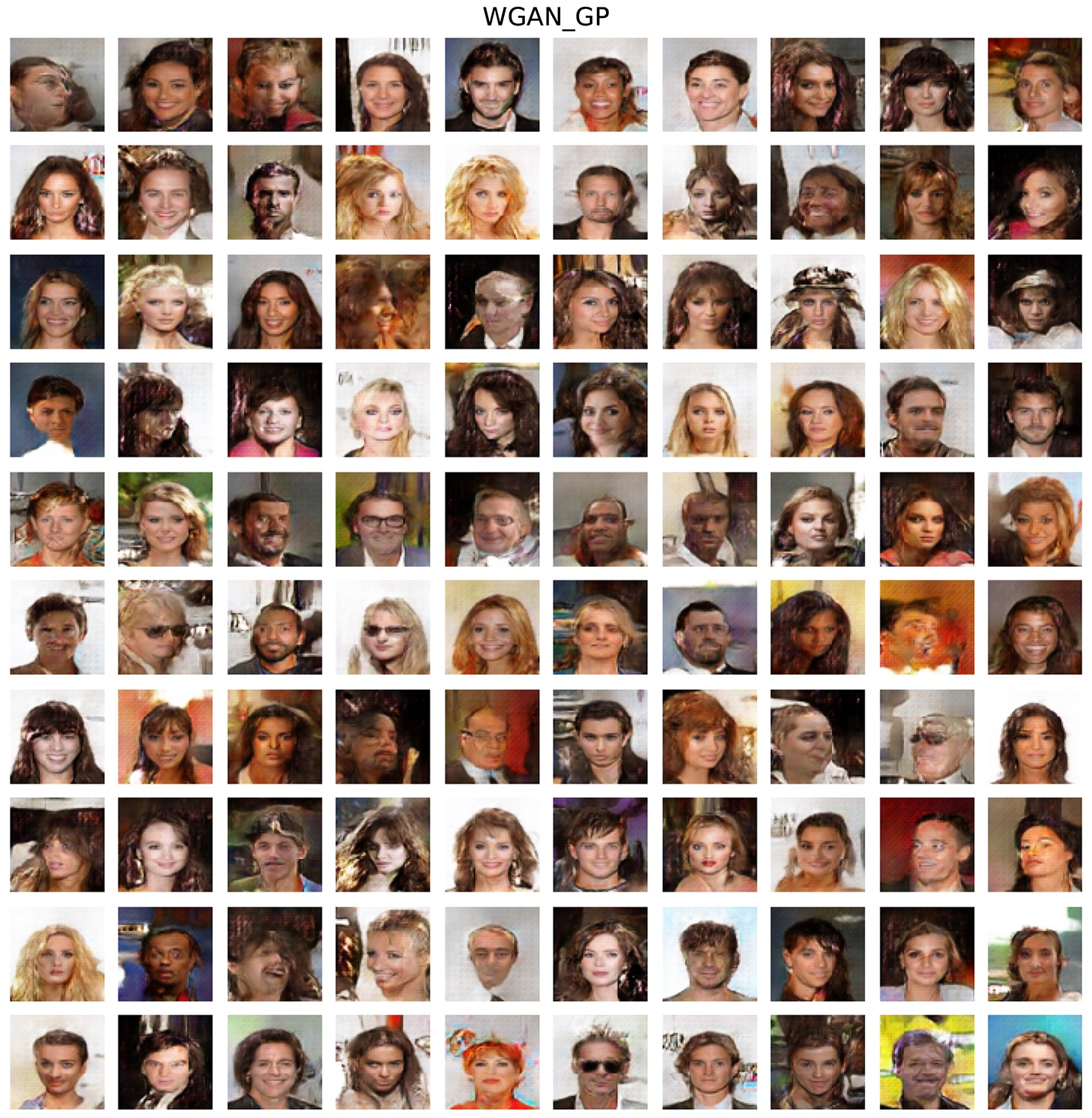}
    }
    \subfigure[WGAN\_GP\_MS]{
        \includegraphics[width=0.33\textwidth]{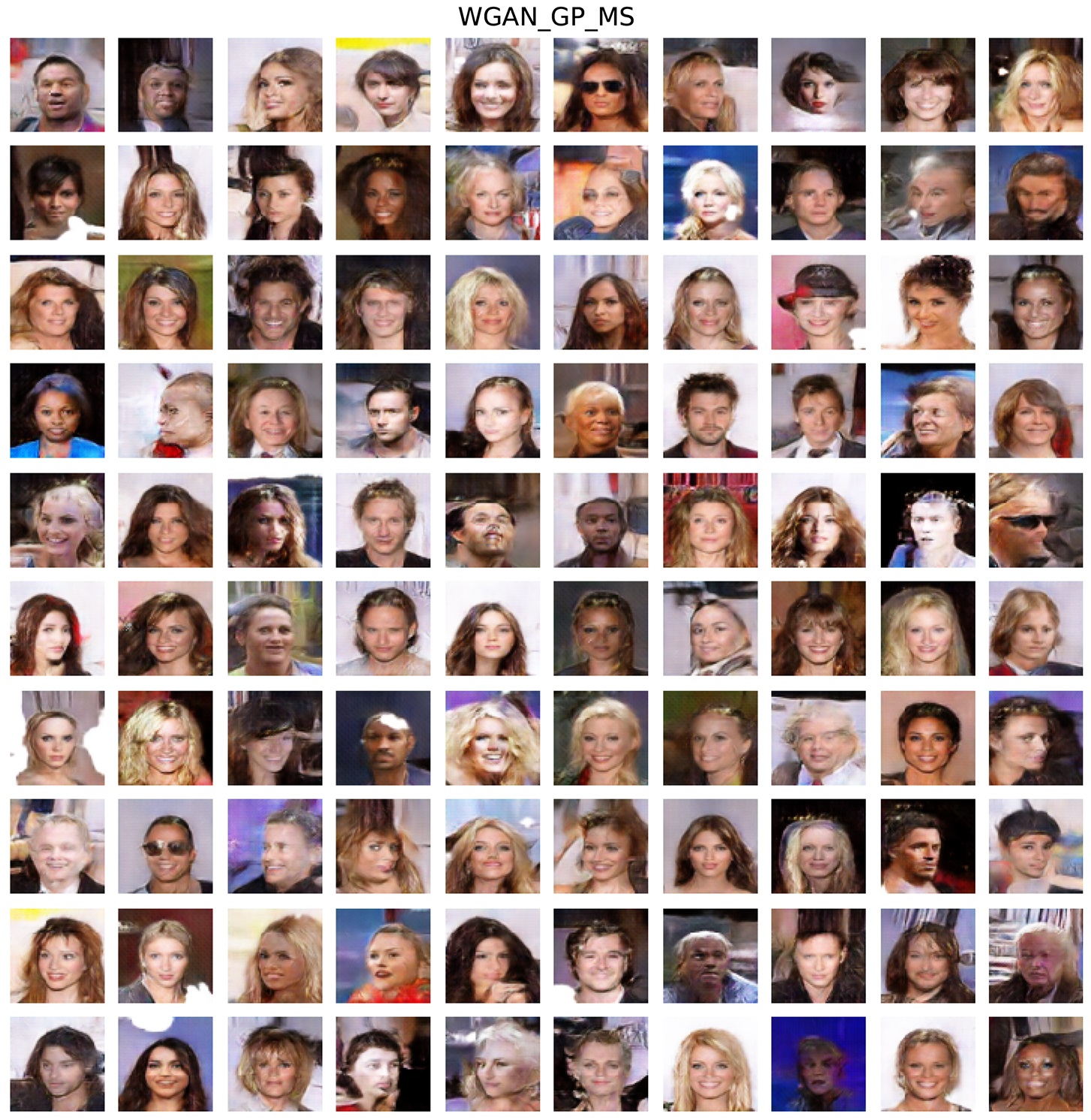}
    }
    \subfigure[WGAN\_GP\_DP with $\lambda_2=10$]{
       \includegraphics[width=0.33\textwidth]{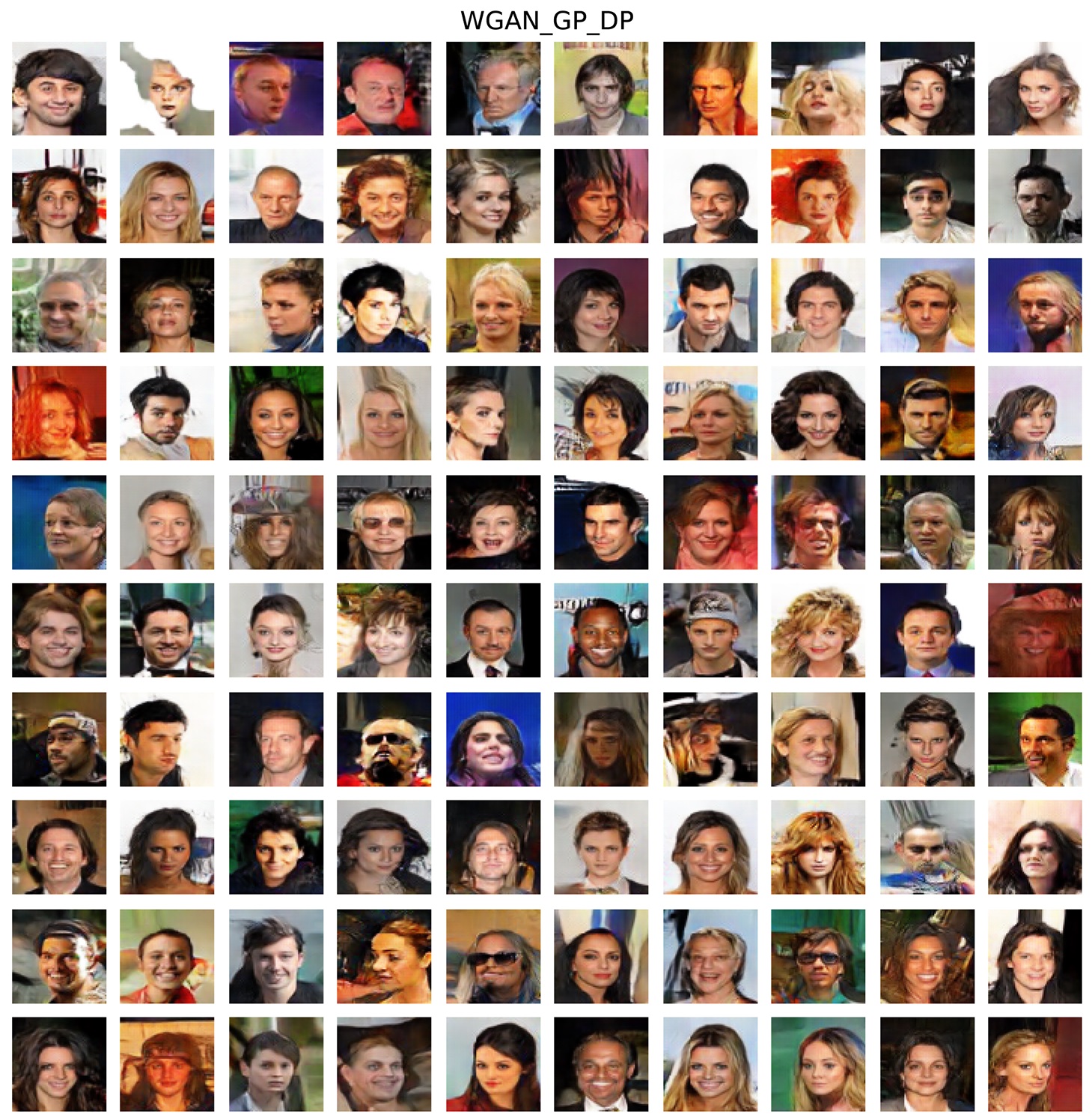}
    }
    \caption{\textbf{Generated Images on CelebA.}}
    \label{fig:celeba_g_imgs}
\end{figure*}

\begin{figure*}[htpb]
	\subfigure[DCGAN]{
        \includegraphics[width=0.31\textwidth]{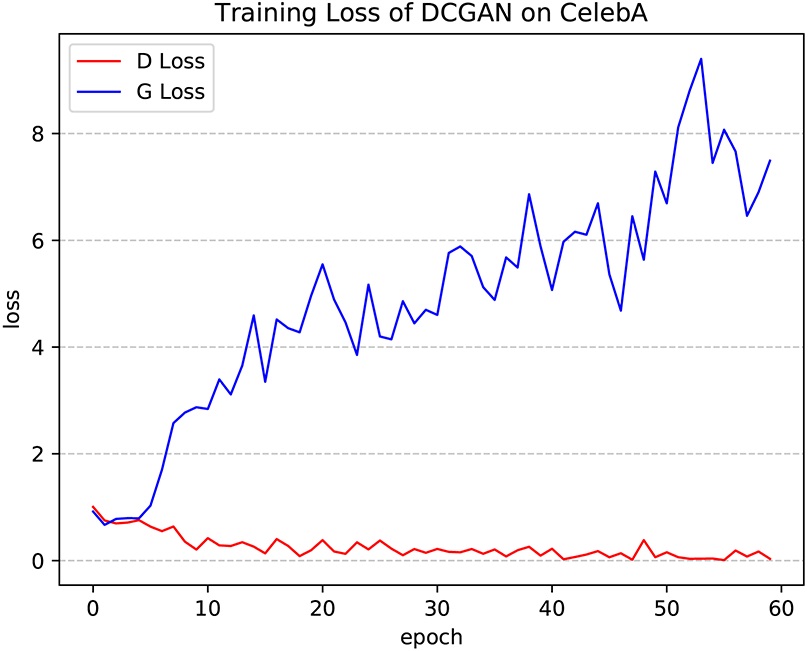}
    }
    \subfigure[DCGAN\_MS]{
        \includegraphics[width=0.31\textwidth]{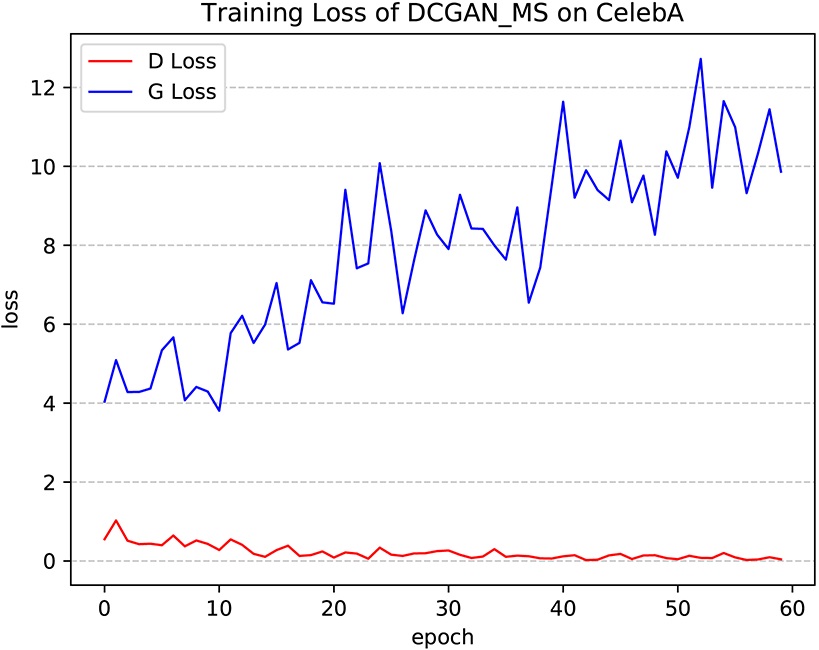}
    }
    \subfigure[DCGAN\_DP]{
       \includegraphics[width=0.31\textwidth]{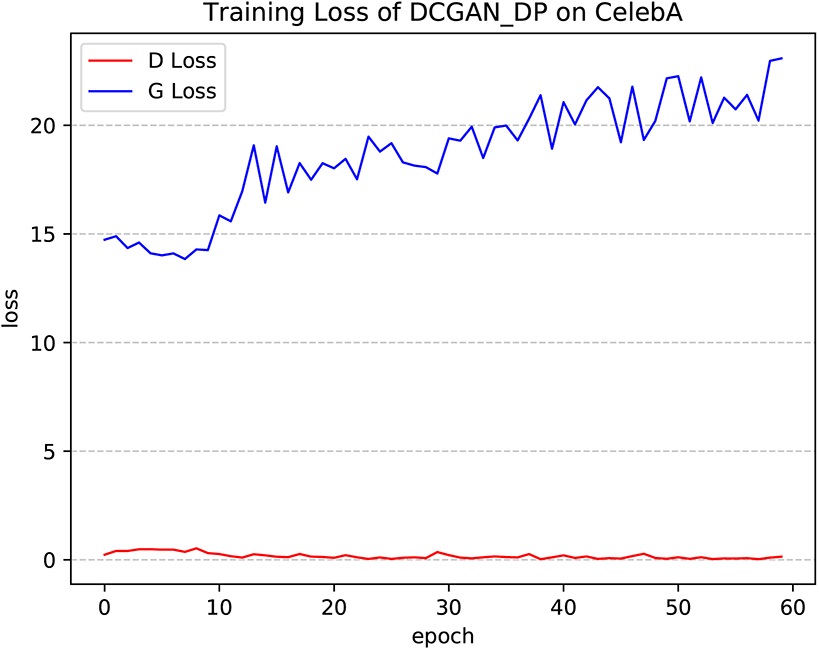}
    }
	\subfigure[WGAN\_GP]{
        \includegraphics[width=0.315\textwidth]{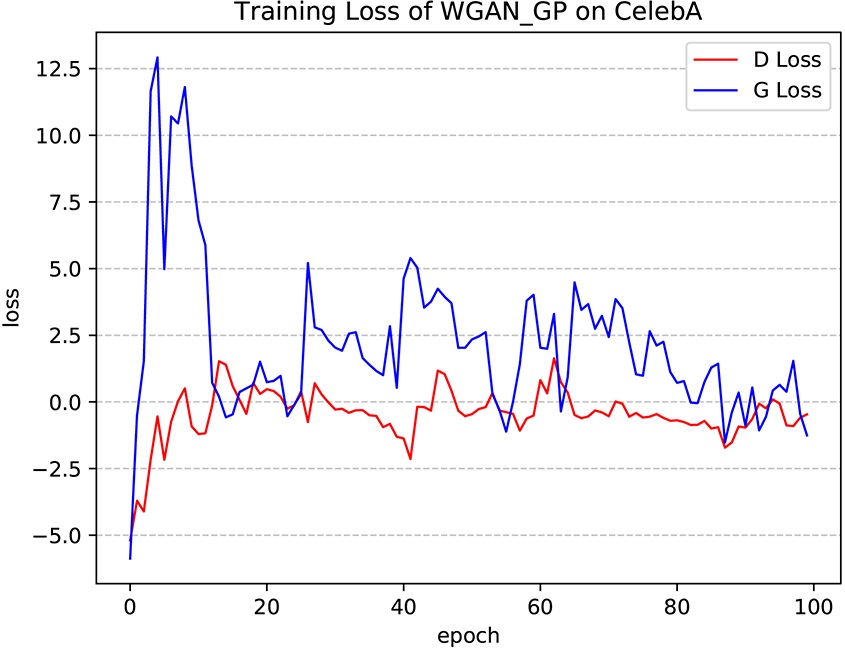}
    }
    \subfigure[WGAN\_GP\_MS]{
        \includegraphics[width=0.315\textwidth]{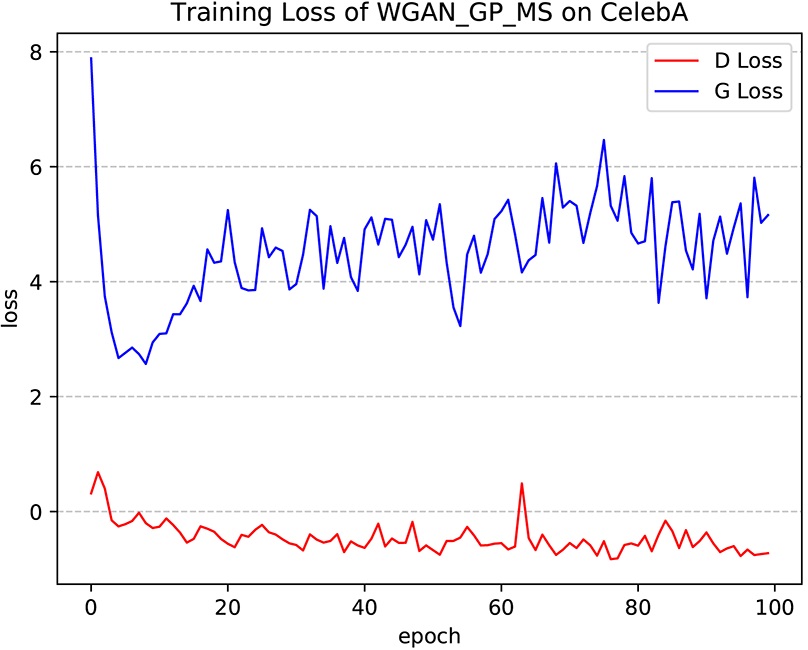}
    }
    \subfigure[WGAN\_GP\_DP]{
       \includegraphics[width=0.315\textwidth]{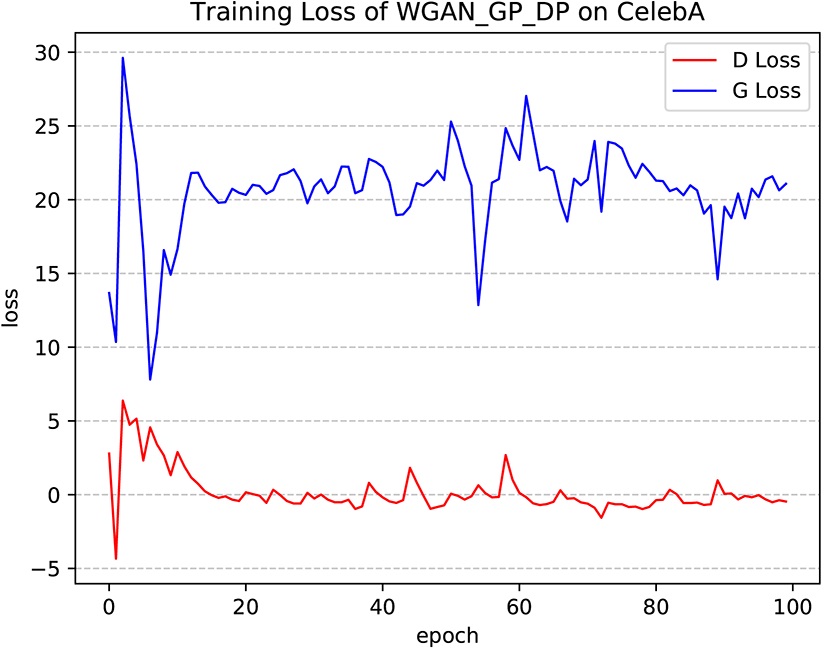}
    }
    \caption{\textbf{Training loss of generator and discriminator on CelebA.} Diversity penalty coefficient $\lambda$ in WGAN\_GP\_DP and DCGAN\_DP is set to 10. $\lambda_{ms}$ in WGAN\_MS and DCGAN\_MS is set to 1.}
    \label{fig:celeba_loss}
\end{figure*}

\begin{figure*}[htpb]
    \subfigure[Our Detecting Framework]{
        \includegraphics[width=0.32\textwidth]{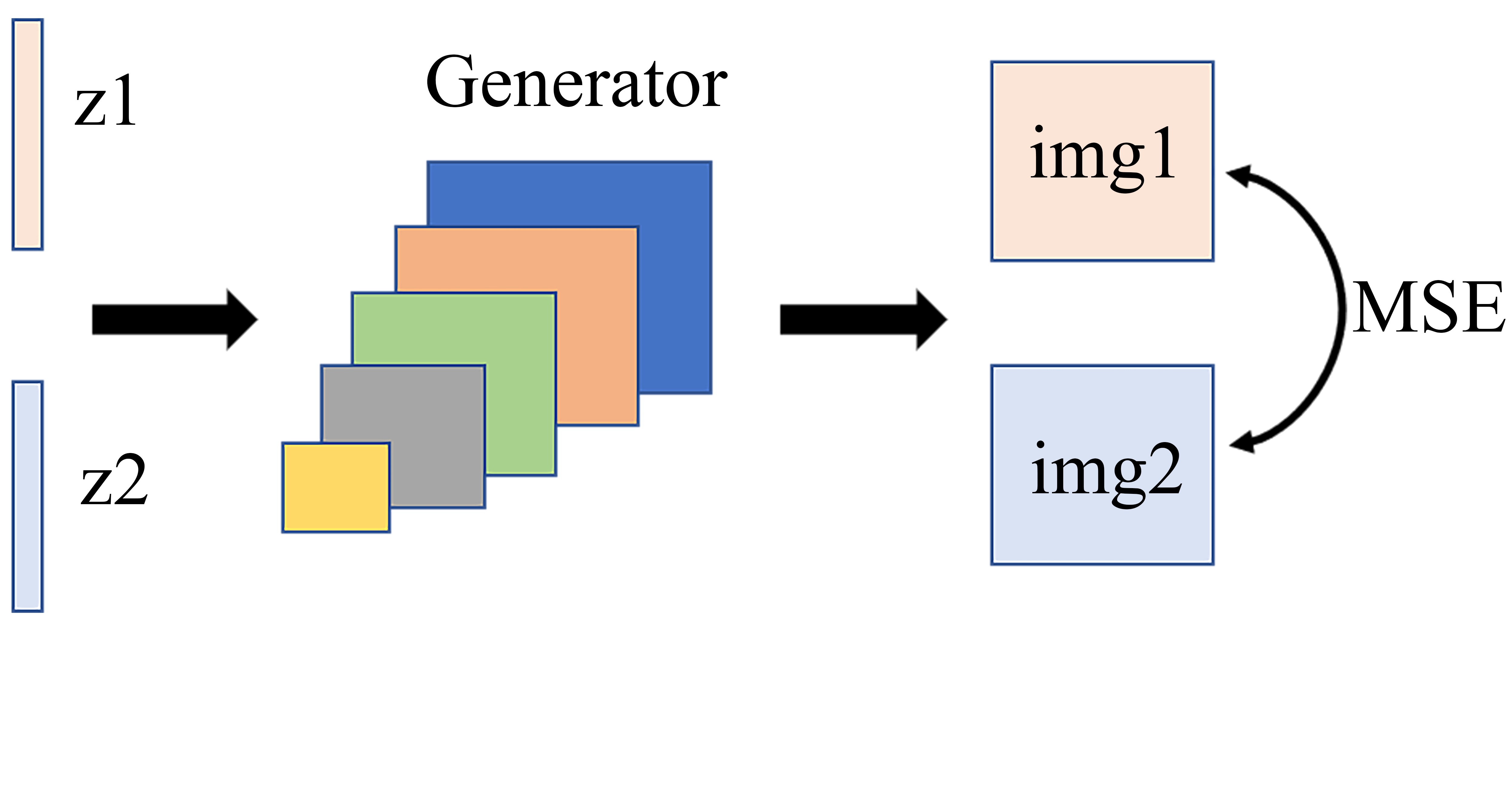}
    }
    \subfigure[Some mode collapse samples on MNIST]{
        \includegraphics[width=0.32\textwidth]{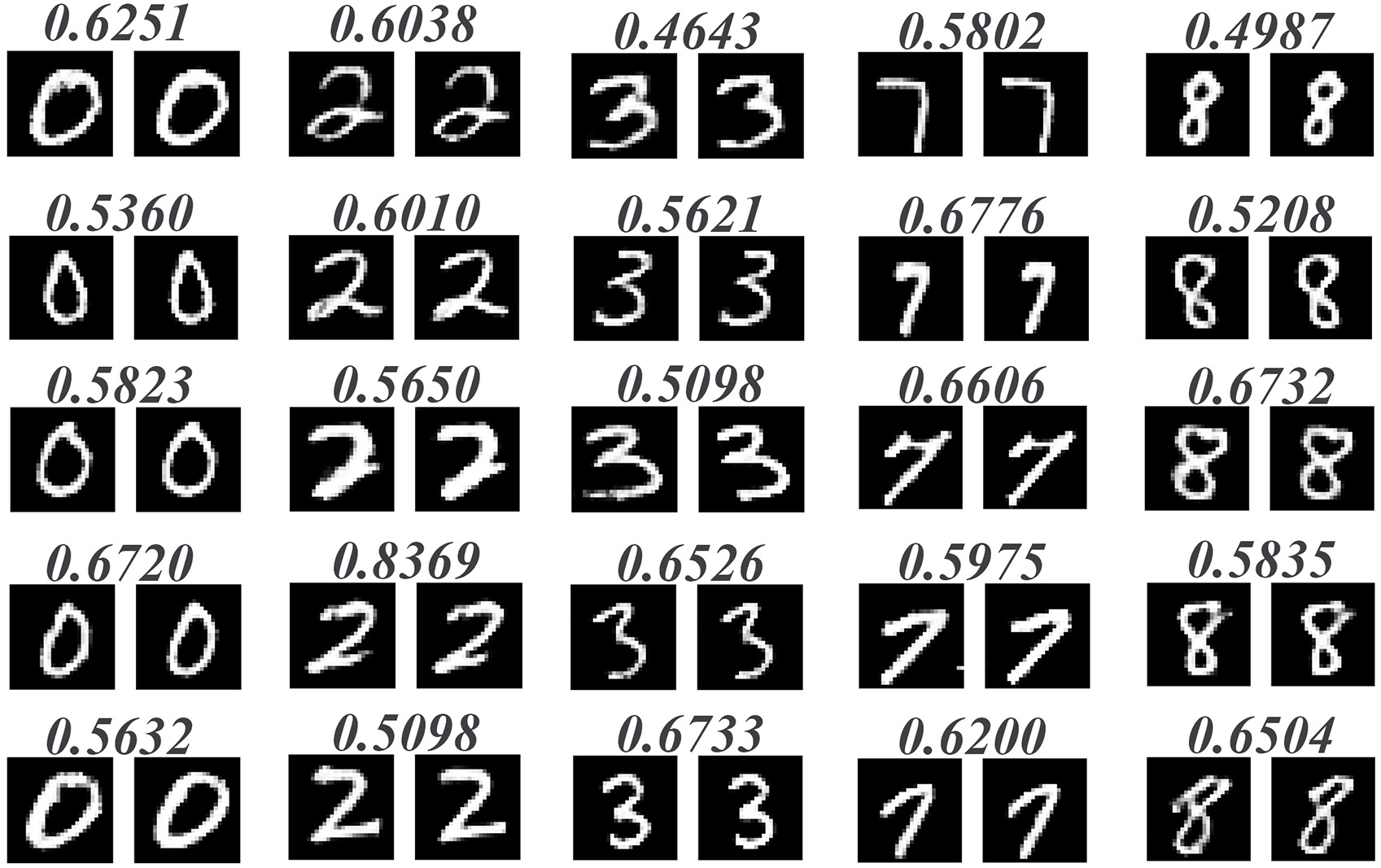}
    }
    \subfigure[Alleviation of mode collapse on CelebA]{
        \includegraphics[width=0.32\textwidth]{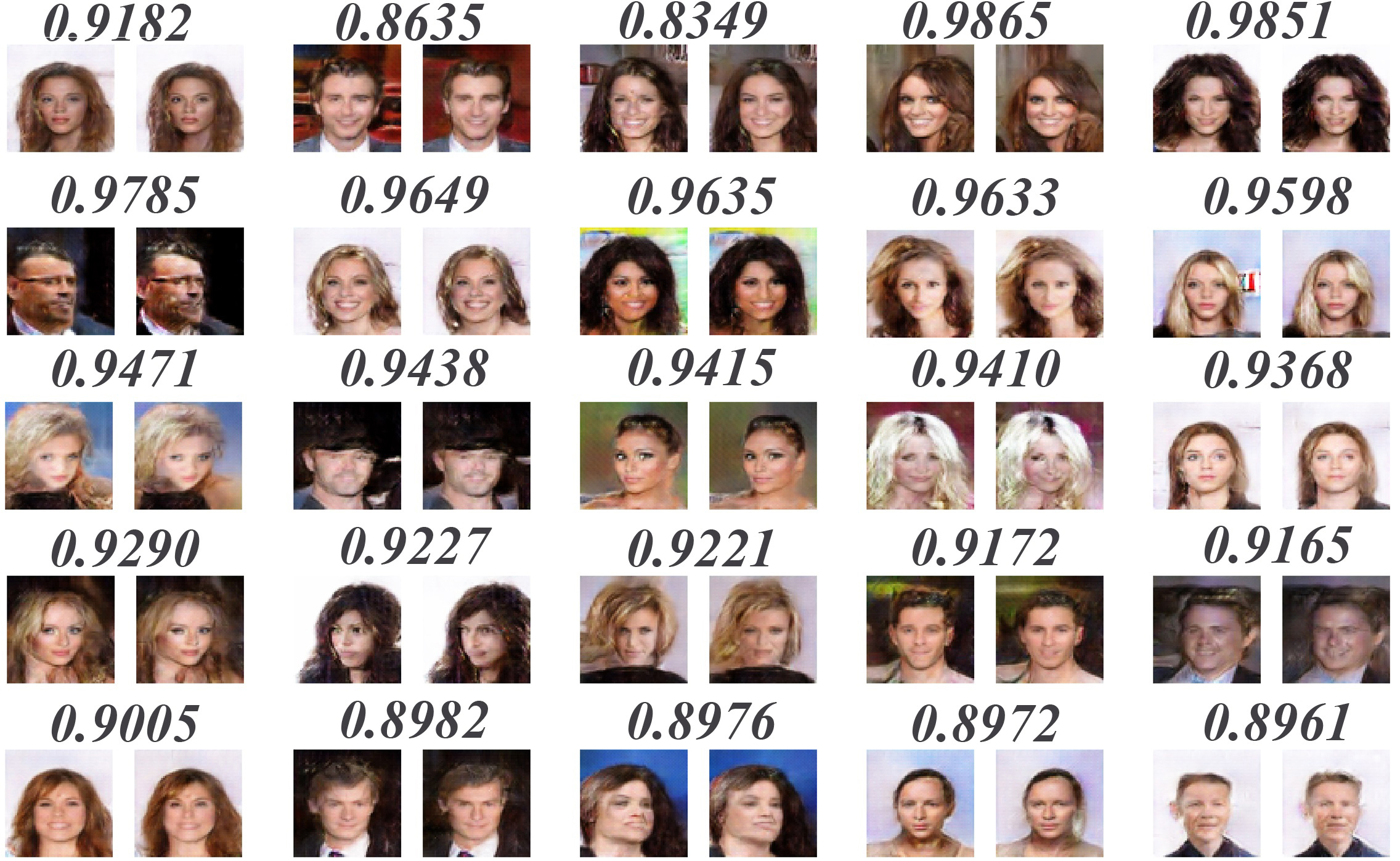}
    }
    \caption{\textbf{Reduction of data diversity detected in our experiments.} We use the adversarial learning method to get two similar images under the MSE metrics. Given random noise $z_1$ and its corresponding generated image img1, we optimize random noise $z_2$ using back propagation to minimize the difference between its corresponding image img2 and img1. Our results show that $z_2$ may be very different from $z_1$, even though their corresponding generated images are similar. (a) WGAN\_GP~\cite{7} on MNIST \cite{12}: some examples of very different noise vectors which are mapped to similar images. The value above each image pair indicates the similarity of their corresponding noise vectors. (b) WGAN\_GP~\cite{7} with \textbf{diversity penalty} on CelebA \cite{23}: results show that the similar images also have corresponding latent vectors with higher similarity compared with (b). Calculation of similarity is shown in formula (\ref{eq:revised-latent-sim}).}
\label{fig:mode-collapse-sample}
\end{figure*}

\begin{figure}[htpb]
\begin{center}
   \includegraphics[width=0.6\linewidth]{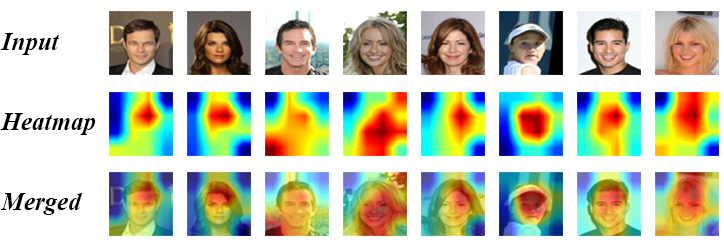}
\end{center}
   \caption{\textbf{Grad-CAM samples on CelebA.} The discriminator of GAN with PDPM can generate facial mask accurately. The architecture of discriminator is shown in Appendix Table \ref{gans-celeba}.}
\label{fig:grad-cam}
\end{figure}

\begin{figure}[htpb]
\begin{center}
   \includegraphics[width=0.5\linewidth]{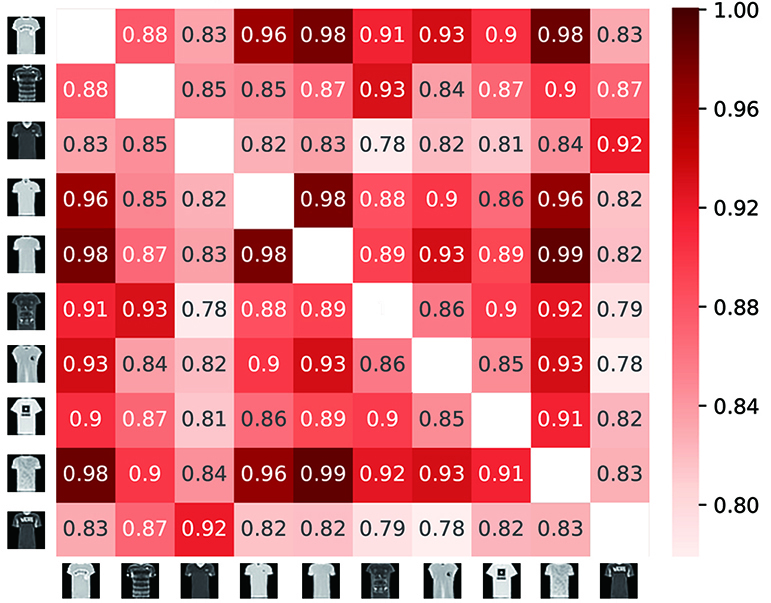}
\end{center}
   \caption{\textbf{Analysis within a Specific Class on Fashion-MNIST.} The result supports that even within a specific class, visually similar images remain close in feature space.}
\label{fig:fm-one-class}
\end{figure}

\end{document}